\documentclass[letterpaper,twocolumn,10pt]{article}
\usepackage{usenix-2020-09}

\usepackage{tikz}
\usepackage{amsmath}
\usepackage{enumitem}

\usepackage{graphicx}  
\usepackage{subcaption}  
\usepackage{algorithm}
\usepackage{algorithmic}
\usepackage{fontawesome}
\usepackage{appendix}

\usepackage{makecell}
\usepackage{multirow}
\usepackage{amssymb}

\usepackage{fdsymbol}

\usepackage{filecontents}
\newcommand{\myparagraph}[1]{\smallskip\noindent{\bf {#1}.}~}

\begin{document}

\date{}

\title{\Large \bf Detecting Dataset Abuse in Fine-Tuning Stable Diffusion Models \\ for Text-to-Image Synthesis}

\author{
{\rm Songrui Wang}*\\
Nanjing University
\and
{\rm Yubo Zhu}*\\
Nanjing University
\and
{\rm Wei Tong}\\
Nanjing University
\and
{\rm Sheng Zhong}\\
Nanjing University
}

\maketitle

\let\oldthefootnote\thefootnote
\let\thefootnote\relax\footnotetext{*Equal contribution}
\let\thefootnote\oldthefootnote

\begin{abstract}
Text-to-image synthesis has become highly popular for generating realistic and stylized images, often requiring fine-tuning generative models with domain-specific datasets for specialized tasks. However, these valuable datasets face risks of unauthorized usage and unapproved sharing, compromising the rights of the owners. In this paper, we address the issue of dataset abuse during the fine-tuning of Stable Diffusion models for text-to-image synthesis. We present a dataset watermarking framework designed to detect unauthorized usage and trace data leaks. The framework employs two key strategies across multiple watermarking schemes and is effective for large-scale dataset authorization. Extensive experiments demonstrate the framework's effectiveness, minimal impact on the dataset (only 2\% of the data required to be modified for high detection accuracy), and ability to trace data leaks. Our results also highlight the robustness and transferability of the framework, proving its practical applicability in detecting dataset abuse.
\end{abstract}

\section{Introduction}
\label{sec:intro}
Text-to-image synthesis, enabled by various generative models such as MidJourney~\cite{midjourney2024}, Stable Diffusion~\cite{rombach2022highresolutionimagesynthesislatent}, and DALL$\cdot$E 2~\cite{gu2022vectorquantizeddiffusionmodel}, has attracted significant attention for its ability to create highly realistic and stylized images. This process involves a model that takes a natural language description as input and generates an image that aligns with the described theme. Among these models, diffusion models like Stable Diffusion have become particularly prominent for text-to-image synthesis. Their powerful capabilities, ease of use, and open-source availability have led to widespread adoption in various applications, attracting substantial commercial investment and benefiting millions of users.

Despite their capabilities, pretrained generative models have limitations in specific synthesis tasks. Typically, fine-tuning these models with domain-specific datasets is necessary to create text-to-image models tailored to particular tasks. For example, fine-tuning Stable Diffusion models with samples from a specific artist can produce images that mimic that artist's style. Similarly, using samples from specific subjects, such as celebrities, politicians, or fictional characters, allows for the creation of models that generate images of those subjects. However, fine-tuning pretrained models for specific text-to-image synthesis requires access to data from specific domains. This data is often a valuable asset to its owner, who has either collected or created it. There is a potential risk that entities using this data for model pretraining might misuse or abuse it.

\begin{figure}[!t]
    \centering
    \includegraphics[width=0.4\textwidth]{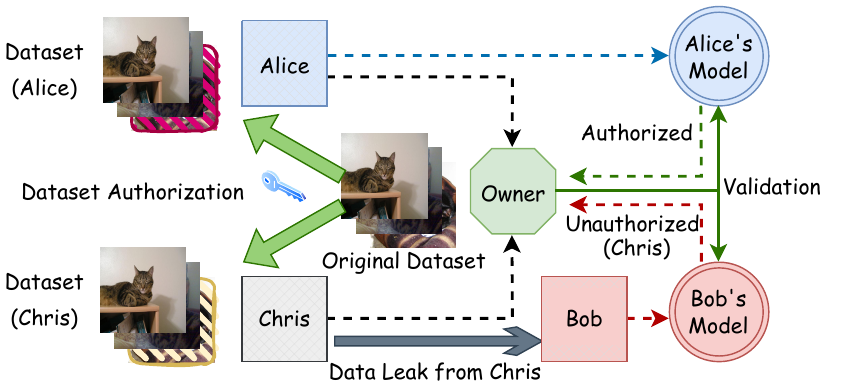}
    \caption{Dataset authorization and potential abuse in fine-tuning text-to-image models. Alice, the data user, requests access to the dataset from the owner. The owner embeds the ownership information and then authorizes Alice to use it. A malicious user may get the dataset without authorization from the owner. The owner should be able to detect any unauthorized use of the data and identify the source of the leak.}
    \label{fig:problem}
\end{figure}

The risk of dataset abuse mainly involves two scenarios. Firstly, authorized entities might use the data to train models for tasks other than those originally intended. For example, a dataset authorized for training generative models for advertising could be misused to train models for creating fake news about specific subjects. Secondly, authorized entities might share the dataset with other parties for profit without proper consent or authorization, violating the rights and interests of the dataset owner. As illustrated in Figure~\ref{fig:problem}, there is potential for dataset abuse, and there is a need for protection. Given these risks, it is crucial to develop defenses against dataset abuse while training text-to-image synthesis models and to protect the copyrights of the datasets involved.
Although previous works, such as \cite{ma2023generative,liang2023adversarial, van2023anti,cui2023ft}, have explored unauthorized image usage in generative models, their approaches cannot apply well to address the dataset abuse problem examined in our paper for several key reasons.  First, large-scale modifications to the dataset may compromise the stealthiness of the protection, making it easier for malicious users to detect and remove the injected watermarks and tokens. Second, although some prior approaches offer effective protection, they do not ensure the harmlessness of the data, rendering it infeasible to train text-to-image synthesis models.  Third, the datasets may be authorized for use by specific entities (authorized data users) for restricted purposes (e.g., limited to specific tasks), but previous works do not adequately prevent or detect such misuse of these datasets, nor can they identify the source of any data leaks.

In this paper, we address the issue of detecting dataset misuse when data is shared with other platforms or users for fine-tuning Stable Diffusion models in text-to-image synthesis. The challenge is dual-faceted. Firstly, once a dataset is shared with proper protection, the original data owner might lose control over it, and it becomes difficult to trace the source of any data leaks. Secondly, implementing heavyweight protection measures or making extensive modifications to the dataset can compromise both its utility and the stealthiness of the protection. Furthermore, there is no guarantee that the ownership information embedded within the dataset will be reflected in the images produced by the text-to-image model after training.

To address the challenges mentioned above, we propose a novel dataset watermarking framework designed to detect potential dataset misuse in text-to-image synthesis with Stable Diffusion models. The proposed framework can achieve the desirable properties: 1) it can accurately determine whether a dataset owned by the data owner has been used to train a text-to-image synthesis model, allowing it to detect dataset misuse and identify the source of a data leak; 2) it only requires modifying a small portion of the data, with the modifications to images and texts being as imperceptible as possible so that the protection cannot be easily detected or bypassed; 3) it does not interfere with the normal use of the data, allowing the trained model to maintain its ability to generate high-quality images.  Furthermore, the proposed framework ensures that the protection is both transferable and robust. Intuitively, our basic idea leverages the unique property of datasets used for text-to-image model training: they consist of image-text pairs.

Our method exploits the dependency between text prompts and images, adopts effective watermarking schemes for images during the text-to-image synthesis model training process, and treats some specific tokens as a backdoor in the model to establish an effective framework. Extensive experiments demonstrate that the proposed framework achieves effectiveness, harmlessness, traceability, and stealthiness. Notably, our approach requires injecting tokens or watermarks into only about 2\% of the data, significantly lower than the requirements of previous methods, to achieve high detection accuracy. Furthermore, we explore the transferability of our method, demonstrating that it works effectively across different datasets and fine-tuning methods. This highlights its practicality, even when the data owner has limited knowledge of how users will utilize the shared dataset.

\myparagraph{Contributions}
To summarize, this paper makes the following major contributions:
\begin{itemize}[leftmargin=*,noitemsep,topsep=0pt,parsep=0pt,partopsep=0pt]
    \item We conduct a thorough investigation of dataset abuse in the fine-tuning of text-to-image models, exploring effective methods for detecting abuse and pinpointing the source of data leaks.
    \item We introduce a dataset watermarking framework, comprised of token and watermark injection, and watermark detection phases. This enables data owners to detect potential dataset abuse by a text-to-image service provider.
    \item We carry out extensive experiments using two datasets and two primary fine-tuning approaches: standard full parameter fine-tuning and Low-Rank Adaptation (LoRA)~\cite{hu2021lora}. These experiments evaluate the effectiveness, harmlessness, traceability, and robustness of our proposed framework. The results further demonstrate the stealthiness of the method, as it requires only imperceptible modifications to a very small portion of the data. Additionally, our framework shows strong transferability across different datasets and fine-tuning methods.
\end{itemize}

\section{Preliminaries}
\subsection{Text-to-Image Synthesis}

A text-to-image model receives a natural language description as input and generates an image corresponding to that description. Since its initial proposal in \cite{mansimov2016generatingimagescaptionsattention}, the field of text-to-image synthesis has expanded rapidly, using prompts in natural language to generate images.

A typical text-to-image model has two major components: a text embedding module and a conditional image generator. Text embedding modules, such as the text encoder of CLIP~\cite{radford2021learningtransferablevisualmodels} or BERT~\cite{devlin2019bertpretrainingdeepbidirectional}, transform textual input into semantic features. Initially, image synthesis primarily utilizes generative adversarial networks (GANs)~\cite{goodfellow2014generativeadversarialnetworks} and variational autoencoders (VAEs)~\cite{kingma2022autoencodingvariationalbayes} as the conditional image generators. However, recent advancements have increasingly employed diffusion models~\cite{gu2022vectorquantizeddiffusionmodel,rombach2022highresolutionimagesynthesislatent,ramesh2022hierarchicaltextconditionalimagegeneration}, due to their ability to generate images of superior quality with intricate details.

\myparagraph{Diffusion Models} The core of a diffusion model is a stochastic differential process known as the diffusion process. This process is split into two main phases: the forward diffusion and the backward denoising. The forward diffusion starts with a distribution of real images $ q(x) $ and progressively adds noise to an initial data sample $ x_0 \sim q(x) $ through a sequence of steps, leading to a series of increasingly noisy images $ x_1, \ldots, x_T $.
As the steps increase, the data sample becomes less recognizable, and at a sufficiently high number of steps, $x_T$ will approximate an isotropic Gaussian distribution. The backward denoising phase is the reverse of the forward process, aiming to reconstruct $ x_t $ from $ x_{t+1} $, effectively generating the original image from a highly noised state. This is feasible with a sufficiently large $ T $, where a true sample from $ q(x) $ can be generated from Gaussian noise $ x_T \sim \mathcal{N}(0, I) $.

A critical aspect of this model is a neural network $ \epsilon_\theta(x_{t+1}, t) $ that estimates the injected noise $ \epsilon $ at each step. The effectiveness of this neural network is crucial, as it is directly responsible for determining the quality of the reconstructed image in the denoising phase. The network is trained to minimize the discrepancy between the actual noise and its predicted value.
The training process ensures that the network learns to accurately predict and compensate for the noise introduced during the forward phase, which is essential for effectively reversing the noise addition to regenerate the original image from its noised state.

\subsection{Generative Model Fine-Tuning}
Training a text-to-image model from scratch is both computationally expensive and time-consuming. The most advanced models are typically trained on datasets containing over five billion images paired with captions.
In contrast, fine-tuning a pre-trained model on a smaller dataset is a more efficient approach to achieving good performance, particularly when the goal is to generate images in specific domains or styles.

Fine-tuning a text-to-image model involves updating the weights of a pre-trained model using a new dataset. We review two major methods for fine-tuning diffusion models: full parameter fine-tuning and Low-Rank Adaptation (LoRA)~\cite{hu2021lora}. The full parameter fine-tuning method updates all model parameters to adapt to new tasks, whereas LoRA only modifies a subset of parameters by introducing low-rank matrices.

\myparagraph{Full Parameter Fine-Tuning} The Stable Diffusion text-to-image model integrates a U-Net architecture with a text encoder to transform textual descriptions into images. The model can be expressed as:
$ x_{\text{generated}} = \mathcal{U}(x_0, \text{Enc}(t); \Theta)$,
where $ x_0 $ is a latent noise vector, $ t $ is the input text, $ \text{Enc} $ represents the text encoding function, and $ \Theta $ denotes the parameters of the U-Net.

Full parameter fine-tuning adjusts the parameters $\Theta$ to minimize the loss between generated and target images, which is defined as:
$L(\Theta) = \frac{1}{N} \sum_{i=1}^N \ell(\mathcal{U}(x_{0,i}, \text{Enc}(t_i); \Theta), x_i)$,
where $ \ell $ is typically a perceptual loss function that quantitatively assesses the differences between the generated image $ x_{\text{generated}} $ and the target image $ x_i $, enhancing both the fidelity and realism of the generated outputs.

\myparagraph{LoRA} The Low-Rank Adaptation (LoRA) approach is initially proposed for Large Language Models (LLMs) by Szegedy et al.~\cite{hu2021lora}. This method aims to address challenges similar to those encountered in DreamBooth, where teaching the model multiple specific concepts necessitated numerous model replicas, highlighting inefficiencies and impracticalities. The basic idea of LoRA is to embed new, small trainable layers, referred to as LoRA layers, into the model such that they do not alter the overall model parameters while maintaining the original model in a frozen state.

When LoRA is integrated into certain or all convolutional layers of U-Net, it involves adjusting the standard convolutional kernel weights $W$ to $W'$, where $W'$ is the sum of the original weights $W$ and an update generated by the low-rank matrices $A$ and $B$: $W' = W + AB$,
where $A \in \mathbb{R}^{k \times r}$ and $B \in \mathbb{R}^{r \times k}$, where $k$ represents the size of the convolution kernels, and $r$, significantly smaller than $k$, denotes the rank, representing the size of the low-rank structure. This approach leverages the efficiency of low-rank matrix factorization to introduce minimal but effective modifications to the pre-trained model, allowing for targeted adjustments without the need for extensive retraining of the entire network.


\section{Problem Statement}
\label{sec:systemgoal}
\subsection{Threat Model}
Two types of entities are involved in the threat model: the defender (i.e., the data owner) and the adversaries (i.e., the data users). We consider datasets of image-text pairs for training text-to-image models. Given an original dataset consisting of image-text pairs, the data owner can determine activation tokens, and generate corresponding watermarked images. The data owner then releases the watermarked dataset to data users, who will utilize the released data to fine-tune their text-to-image models, thereby enhancing their overall performance in terms of comprehensive or generation capabilities on specific subjects. Adversarial behaviors refer to the unauthorized usage or unapproved sharing of the dataset.

\myparagraph{Data Owners} Data owners could be entities that generate, collect, and label image pairs, such as artwork styles and specific items. By authorizing specific third parties to use their datasets of image-text pairs, the data owners can unleash the value of the data in AI-powered services and profit from this authorization. The data owners intend to generate watermarks and inject them into the original datasets of image-text pairs. Furthermore, the data owners not only want to know if their data has been misused but also to have the ability to trace the source of any data leaks. We assume that the data owners can freely modify their data on images and texts. We also assume that the data owners can use the inference services powered by the models trained by the data users. We first consider the case where the data owners know the specific model and the fine-tuning method used by the data users. We also consider the black-box setting where the data owners have no such knowledge about the model or training process.

\myparagraph{Data Users} Data users train or fine-tune a text-to-image model to meet certain needs with the data provided by the data owners.  Data users have full control over how to fine-tune their text-to-image model with the data (e.g., using LoRA~\cite{gandikota2023conceptslidersloraadaptors}, or standard training~\cite{platen2022diffusers}). A benign data user obtains authorization from the data owner and only uses the data for a specific, authorized task. This means the benign data user will not use the data for any other unauthorized tasks or share the data with others without permission. Specifically, an adversary could be the authorized user who shares the data without permission. Or, the adversary may be a data user who obtains data from an authorized user who has no right to share the data, instead of getting authorization from the data owner.
In addition, we also consider the case where an adversarial data user may gain access to the data based on leaked information from other services that have been authorized to access the data. We refer to a data user who leaks the data (intentionally or unintentionally) as a \textit{data leaker}, and a data user who uses the data without authorization as a \textit{data abuser}.

We also assume that an adversary may attempt to bypass the protection using techniques such as compression, sharpness enhancement, noise addition, blurring, and resizing to preprocess the data. However, we consider \textbf{it unreasonable for an adversary to delete all the text or images in the dataset} to evade detection, as the image-text dataset is valuable, and such actions would make the text-to-image synthesis task impossible, thus undermining the adversarial goal of misusing the data.

\subsection{Design Goals}
Our system is targeted at detecting unauthorized data usage and identifying the leaker, while the regular authorized usage of the data by data users is not affected. To achieve the above goals, we hope to build a system with the following properties:

\begin{itemize}[leftmargin=*,noitemsep,topsep=0pt,parsep=0pt,partopsep=0pt]
    \item \textbf{Effectiveness:} To detect unauthorized use of the data, a data owner watermarks image-text pairs in the dataset before sharing them with data users. Subsequently, the data owner can detect whether the data has been used by performing queries on the trained model and checking the outputs. It is essential to achieve high detection accuracy in watermark verification in order to effectively detect data misuse.
    \item \textbf{Harmlessness:} The injected watermarks should not affect the normal use of the data. Authorized data users can still utilize the watermarked dataset to train or fine-tune their text-to-image models. The trained models should maintain a high quality of generated images for regular prompts used as inputs in the services provided by the data users.
    \item \textbf{Traceability:} Data provenance is crucial for data sharing. When the data owner detects that an unauthorized model is using the watermarked dataset, the system should also provide the capability to identify the source of the data leak or the party that illegitimately reauthorizes the data.
    \item \textbf{Stealthiness:} All changes made by the system to the data should be stealthy. First, the proportion of the dataset that undergoes modification should be kept small. The stealthiness of the watermarks is crucial to prevent detection and filtering by unauthorized data users.
    Second, the degree of modification applied to a selected image-text pair should be small, trying to make the modifications imperceptible.
\end{itemize}

\subsection{Straightforward Approaches: Infeasibility}
We consider two intuitive straightforward approaches to address our problem.

The first approach involves applying watermarks to a significant number of images in the dataset, ensuring that the watermark remains effective even after undergoing the diffusion model’s training and generation processes. However, this approach has two main limitations. Firstly, to maintain the watermark's effectiveness, the majority of the images in the training data would need to be altered. Such a high rate of watermark injection could potentially disrupt the normal functionality of the model. Secondly, this method makes it impossible to trace the source of a data leak, as the watermark appears identical across all instances.

The second approach involves using techniques that embed information within the data, allowing the dataset to be tracked for different users with the hidden information. To investigate the feasibility of such techniques, we utilize HiDDeN~\cite{zhu2018hiddenhidingdatadeep}, which assigns unique keys to different users and embeds these key messages into the original images. This method enables the detection of hidden messages within the protected dataset in its original form. However, when these images are processed through the diffusion model, the key messages do not remain intact, making it nearly impossible to detect the keys in the generated images. Using an open-source implementation~\cite{khachatryan2018hidden} of HiDDeN, we have embedded binary strings of length eight as key messages. In our exploratory experiments, the results show a bitwise error rate of $0.08$ on the original protected data. After fine-tuning the diffusion model with this data, the bitwise error rate increases to $0.43$ (with $0.5$ representing a random baseline) on the generated images. This indicates that the hidden messages are significantly distorted and ineffective during the diffusion model's training and generation stages.

In summary, two main factors cause these straightforward approaches to fail. First, there is an inherent conflict between the effectiveness of protection and its harmlessness. Second, these straightforward solutions, which focus solely on the image data, are insufficient for achieving traceability to identify the source of a data leak.

%
%

\begin{figure*}[!t]
    \centering
    \small
    \includegraphics[width=0.71\textwidth]{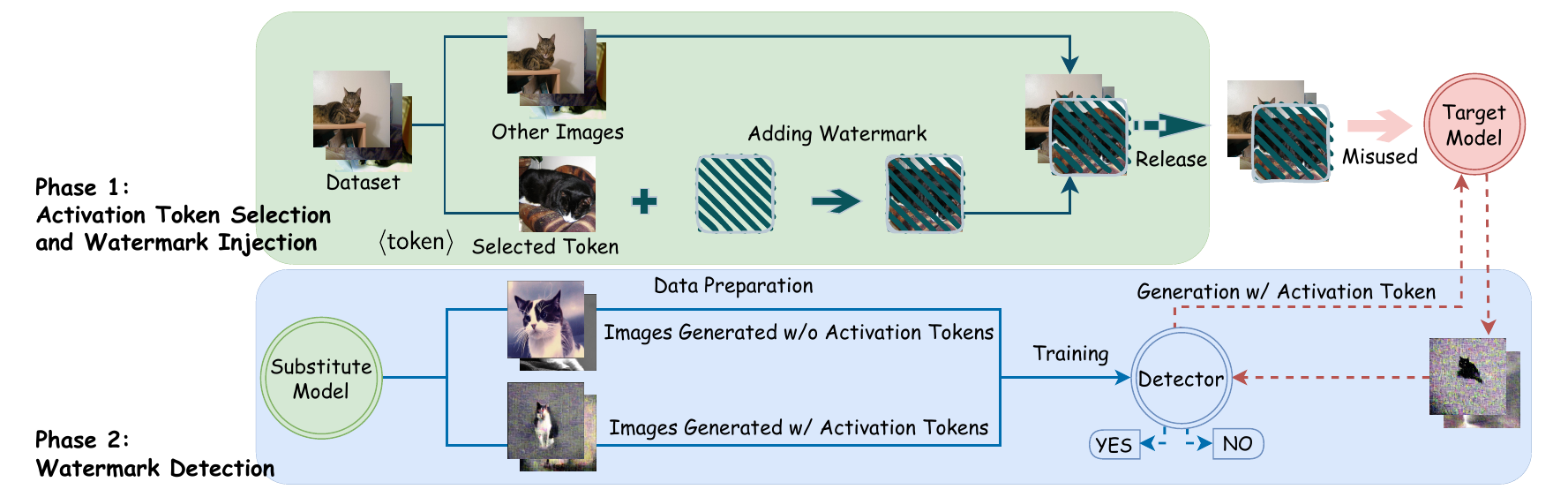}
    \caption{The proposed dataset watermarking framework.}
    \label{fig:framework}
\end{figure*}

\section{Framework}
\label{sec:Framework}
To address the challenges mentioned above, we adopt the concept of the backdoor. In our problem, a backdoor can be considered a common feature present in a small subset of images. By applying a watermark to this small subset, we maintain a low rate of watermark injection while ensuring the effectiveness of the protection. For text-to-image diffusion models, it is intuitive to use specific tokens as the trigger of the backdoor. We can either slightly modify the text in the original data or leave it unchanged while selecting specific tokens that correspond to certain text-image pairs. The watermark is then embedded into these images, and the selected token acts as the trigger that can activate the watermark feature. This approach enables us to detect suspicious models effectively. Below we first present the overview and then detail the major components of the proposed framework.

\subsection{Overview}
Our framework consists of two phases: \textit{{activation token selection and watermark injection}} and \textit{{watermark detection}}:

\myparagraph{Activation Token Selection and Watermark Injection} Our primary idea is to use text tokens instead of image watermarking patterns to identify different data users. It is nearly infeasible to choose distinct and easily distinguishable image perturbations to represent user identities due to the continuous property of the image perturbations. In contrast, the text tokens offer ample specificity for us to leverage, allowing us to leverage them effectively. Moreover, combining multiple text tokens can significantly expand the pool of activation tokens simply and efficiently. We select unique tokens to represent a user's identity in the text data while creating effective watermarks for the associated images. This approach ensures clear differentiation between users and enables us to trace the source of data leaks in practice.

\myparagraph{Watermark Detection} The basic idea is to develop a discriminator that can detect watermarking patterns in images generated by text-to-image models for the data owner (verifier). Our approach allows the data owner to use activation tokens from data users to prompt the suspected model, generating images that contain watermark features. Given a user's activation token, the data owner prompts the text-to-image model to generate images and then uses the discriminator to detect watermarking patterns in these images. If the watermarking patterns are detected with high accuracy, the data owner can determine that the dataset used by the suspected model is leaked from the user represented by the activation token. This process confirms whether the model has utilized the data owner's dataset and helps identify the source of the dataset leak through the binding between activation tokens and data users. The higher the detection accuracy, the greater the success rate of identifying the source of a data leak. Therefore, our primary goal in the design is to achieve high detection accuracy.

Figure~\ref{fig:framework} illustrates the overall framework of our method. We will detail the strategies adopted in the activation token and watermark injection phase in Section~\ref{Injectwatermarkandtoken} and the method of training the detector in Section~\ref{discriminator}. The watermarking schemes and activation token selection details will be provided in Section~\ref{sec:details}.



\subsection{Activation Token Selection and Watermark Injection}
\label{Injectwatermarkandtoken}
This process stems from a key observation during the fine-tuning of text-to-image models, where the model is trained to align text with corresponding images. This alignment allows our framework to embed information into both texts and images, leveraging their dependencies to create activation tokens for different data users and embed watermarks. Our method exploits the alignment between the diffusion model and CLIP during training. This alignment establishes a one-to-one mapping between tokens and image features. We select a unique token for each user and embed watermarks into the images associated with this token in the dataset.
This approach is effective because the model learns to generate the watermark feature when the specified token appears in the input text.

%

\myparagraph{Alignment Strategies} We explore potential strategies for aligning tokens in text prompts with embedded watermarks in images:

\textit{Token-Watermark Alignment}: This strategy allows the data owner to embed specific tokens into text and apply watermarks to images to align the tokens with the watermarks. This alignment enables the tokens to activate the watermarks, facilitating the detection of the watermark by the data owner when needed. By modifying both text and image parts in a dataset pair, this approach is identified as the most straightforward and effective among the four strategies examined.

\textit{Watermark-Adding Alignment}: In this strategy, watermarks are applied solely to images, while pre-existing tokens from the text are selected to align with these watermarked images. For example, one could choose tokens that appear frequently in the text or those that hold particular significance. This alignment ensures that the watermark can be activated by the selected token when detecting the watermark. Since this strategy does not require the injection of new tokens into the text, it maintains a high level of stealthiness.

\textit{Token-Injection Alignment}: This approach involves the addition of new tokens to the text without modifying images. The key concept here is to align these injected tokens with existing patterns in images, enabling the tokens to activate these features and thereby allowing the data owner to detect the use of the dataset. The main challenge in implementing this strategy is identifying which image features can be effectively activated by the tokens, which may limit its applicability across diverse datasets. This approach does not require modifications to the images, which enhances its stealthiness.

\textit{Natural Alignment}: This strategy focuses on identifying natural relationships between existing tokens in text and patterns in images without making any modifications. The goal is to establish a sufficiently large set of such relationships to enable the creation of distinct token sets for different data users. While this method is highly stealthy, its practicality may be limited in real-world scenarios due to the difficulty in finding such natural alignments across various datasets.

The last strategy, Natural Alignment, is the most stealthy but also the most challenging to implement in practical scenarios involving real-world datasets.  In addition, this strategy cannot support the data owner in identifying the source of the data leak. The Watermark-Adding and Token-Injection strategies are comparable in the extent of dataset modifications they require. However, empirical evidence~\cite{DBLP:conf/ndss/LiJDLW19} suggests that it is easier and more stealthy to modify images than text. This is because perturbations in images can be made virtually imperceptible, whereas changes in text are typically obvious.

Therefore, we have selected the Token-Watermark Alignment (TWA) and Watermark-Adding Alignment (WAA) strategies to realize our design for implementing the watermark injection phase. We describe the details of them as follows. Denoted by $A = (A_{\text{image}}, A_\text{text})$ the dateset to be protected, where $A_\text{image}$ represents the image data and $A_\text{text}$ represents the text data. Once a data user requests the dataset from the data owner, the data owner executes the activation token and watermark injection, which includes the following components:

\myparagraph{Activation Token Selection} For the $i$-th data user, the framework selects the activation token $T_i$ to represent its identity. In the Token-Watermark Alignment method, the activation token can be an arbitrary word token $T_i$ generated by the data owner. They can also be an out-of-vocabulary (OOV) token. In the Watermark-Adding Alignment method, the activation token is selected from existing tokens within the dataset. A common method for choosing the word tokens that make up the activation token involves analyzing the frequency of words in the dataset.

\myparagraph{Watermark Injection} The data owner embeds watermarks directly into images, treating them as perturbations rather than using traditional bounding box watermarks. Denoted by $W_i(x)$ the watermark scheme applied for the $i$-th data user on an image $x$. In \textit{Token-Watermark Alignment}, the data owner randomly selects $M$ image-text pairs from the dataset. For each pair, the activation token is inserted at the beginning of the text, and the watermark is embedded into the corresponding image. In the Watermark-Adding Alignment method, the data owner identifies image-text pairs that already contain the chosen activation token in the text and embeds the watermarks into the corresponding images.

Based on the above two steps, we have modified the dataset $A$ by injecting the activation token $T_i$ (if we choose the Watermark-Adding Alignment method, $T_i$ remains the pre-existing words in the dataset) and embedding the corresponding watermarks into the image data $A_{\text{image}}$ by $W_i(\cdot)$. We denote the dataset that includes the modified image-text pairs as $A_i$ and send it to the data user. The data user can then use this data to train its model. As stated in the design goal, the user's model should function normally after fine-tuning. Denote the number of the whole dataset as $N$, and the number of image-text pairs that need to be modified by watermarks and the activation token as $M$. We call $p=\frac{|M|}{|N|}$ the injection ratio.

\subsection{Watermark Detection}
\label{discriminator}
During the watermark detection phase, the data owner utilizes a discriminator (detector) and an activation token $T_i$ to verify whether the suspected model has used the dataset authorized to the $i$-th data user. Based on the design goals, we not only need a binary classifier to detect whether the model used our protected data but also need to track who leaked the data. We regard the output of the suspected model as the input for the discriminator. Specifically, the watermark detection phase consists of two major parts: training the detector and activating the suspected model for detection.

\myparagraph{Detector Training}
Our goal is to train a discriminator $D$ that can detect whether the output images generated by a suspected text-to-image model contain the watermark features. More specifically, we want to train a unified detector that can capture the features of watermarks for all data users instead of training a separate detector for each user. The first step is to generate a large number of images with or without watermarks to construct the dataset for training the discriminator. Denote the suspected model by $\mathcal{S}$. We first train a substitute model, denoted by $\mathcal{S}^\prime$, which is fine-tuned on a dataset with modified image-text pairs, similar to a model trained by a data user.

For a straightforward detection training method, We could first examine a scenario with a single data user, employing a straightforward method to construct the training set. Using the activation token to trigger $\mathcal{S}^\prime$ and output images with watermark features, we denote the set of these images as $I_{1}$, assigning them labels of $1$. Conversely, normal prompts without activation tokens produce a set of images denoted as $I_{0}$, which we label as $0$. Then we could train a binary classifier with these labeled data.

However, this approach is limited and flawed in cases involving multiple users. Since each user has her/his own activation token,  testing a suspicious model with all activation tokens and their corresponding classifiers, one by one, becomes cumbersome and inefficient for large-scale sharing. Furthermore, it is possible that a user’s activation token could already have a specific meaning, such as ``car'', which may lead the detector to classify based on the features of the car in the image rather than the presence or absence of watermarks.

To tackle these challenges, it is necessary to develop a more generalized detector training method. For data distributed to different users using the same token-watermark alignment method, it is expected that images generated by the trained model, when prompted with the activation token, will exhibit consistent watermark features. This is the pattern we aim for the classifier to learn. We approach this by simultaneously considering a set of target models. The images produced by each target model using its designated activation token are \textbf{combined} as $I_{1}$, while images generated by each model with normal prompts are \textbf{grouped} as $I_{0}$.  In this manner, we only need to train a single classifier using $I_{0}$ and $I_{1}$ to effectively detect the watermark feature.

We then use the labeled dataset, composed of $I_{1}$ and $I_{0}$ with their corresponding labels, to train the detector $D$. We employ binary cross-entropy loss to train the discriminator, aiming to distinguish whether the output image contains the watermark feature. The loss function is defined as:
\[
L_{D} = -\frac{1}{N}\sum_{i=1}^{N}y_{i}\log(D(x_{i}))+ (1-y_{i})\log(1-D(x_{i})),
\]
where $N$ is the number of training data, $y_{i}$ is the label of the $i_{th}$ data, and $D(x_{i})$ is the output of the discriminator.

We note that although a substitute model is required to generate data for training the detector, the feasibility and practicality of this approach can be justified in three ways. First, the dataset is owned by the data owner, who can utilize it to train the substitute model. Second, only a single unified detector needs to be trained for all data users, simplifying the process since both the substitute model and the detector only need to be trained once. Third, the transferability of the proposed method (as demonstrated by the experimental results presented in the evaluation) shows that a detector trained on a specific dataset or with a specific training method can be effectively applied to other cases with different datasets and training methods.

\myparagraph{Activation and Detection} In this phase, we can freely choose tokens as the input prompt to test the suspected model. We enumerate all the tokens distributed to data users and use the discriminator $D$ to check if the output image contains the watermark feature. This allows us to determine which activation token has been used by the suspected model. Given the one-to-one correspondence between the activation token and the data user, we can track who leaked the data. The data leaker will be held responsible for the data abuse.

\section{Design Details}
\label{sec:details}
As described in Section~\ref{sec:Framework}, the data owner should select a small portion of image-text pairs in the dataset for injecting/selecting the activation token and adding the watermark. After fine-tuning the diffusion models with this dataset, the activation token could work as the prompt to trigger the model to generate the image with the watermark. This section will introduce the image watermarking schemes and activation token selection methods. Section~\ref{sec:watermarkingSelect} details the design of suitable watermarking for target images. Section~\ref{sec:tokenSelect} describes the process of designing an appropriate activation token. Finally, Section~\ref{sec:LargePub} explores how to establish a suitable activation token when a dataset is accessed by a large number of users.

\subsection{Image Watermarking Schemes}
\label{sec:watermarkingSelect}
To watermark the dataset for verification, we must carefully select a watermarking scheme for specific images in the dataset. This process presents two major challenges. The first challenge is achieving effective watermarking without significantly harming the text-to-image tasks for regular prompts. We can only embed watermarks in a small portion of the dataset, and the scale of the watermarks must be reasonable. The second challenge is ensuring the watermarks remain stealthy while maintaining their effectiveness. If the watermarks are easily detectable, unauthorized data users could identify and remove the watermarked images.

\begin{figure}[tb]
  \centering
  \begin{subfigure}[b]{0.35\textwidth}
       \includegraphics[width=\textwidth]{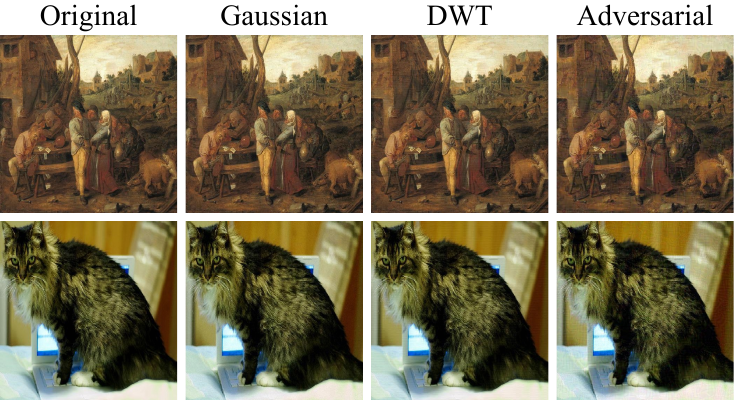}
   \end{subfigure}
   \caption{Examples from the WikiArt and COCO datasets before and after watermarking.}
   \label{fig:Example-1}
\end{figure}

To tackle the challenges outlined, we consider three settings based on different background knowledge concerning the subjects (i.e., images) for embedding the watermarks and the targets (i.e., the models to be trained on the dataset) for generating the watermarks. We have explored three concrete watermarking methodologies for these settings, respectively. The effects of these three watermark types are shown in Figure~\ref{fig:Example-1}, where they appear almost invisible to the human eye, demonstrating the stealthiness of the watermark injection.

\myparagraph{Image-independent Watermarking} In the first setting, the watermarking scheme is not dependent on the target model that the data user has nor the images that need to be embedded in the watermarks. The data owner only needs to know the format (e.g., size, color channels) of the image to embed the watermark. We choose Gaussian noise in this setting as the watermarks.  Due to the space limit, we put the details of the Gaussian watermarking in Appendix~\ref{app:watermarking_methods_gaussian}.

\myparagraph{Target-independent Watermarking} The second setting assumes that the watermarking scheme does not depend on the target model, but rather generates watermarks based on the content of each image. We utilize the Discrete Wavelet Transform (DWT) watermarking~\cite{xia1998wavelet} for this setting. DWT is a versatile technique that decomposes an image into multiple frequency components, allowing for the precise placement of watermarks in areas that minimally affect visual quality. Due to the space limit, we put the details of the DWT watermarking in Appendix~\ref{app:watermarking_methods_dwt}.

\myparagraph{Adversarial Watermarking} In this setting, we assume that the data owner has access to both the image data and the target model or a substitute model. We leverage the technique from the adversarial example to generate the watermarks in this setting. The concept of adversarial example is proposed by Szegedy et al.~\cite{szegedy2013intriguing}. Various methods exist for creating these adversarial examples, such as the Fast Gradient Sign Method (FGSM)~\cite{goodfellow2014explaining} and the basic iterative method~\cite{kurakin2016adversarial}. In our approach, for diffusion models with parameter $\theta$ and an original image $x$, we seek to create an adversarial example $x' = x + \delta$ by introducing adversarial noise $\delta$. Formally, we employ the following formulation to approximate the process of generating adversarial examples:
\begin{align*}
    \delta & = \arg\max_{\delta} \mathcal{L}(\theta^*, x + \delta)\\
    \text{s.t.}  \quad \theta^*&  = \arg\min_{\theta}\mathcal{L}(\theta, x) \land  \|\delta\|_p \leq \eta,
\end{align*}
where $\mathcal{L}(\theta, x)$ denotes the loss with respect to the parameters $\theta$ and the image $x$. As the Fully-trained Surrogate Model Guidance (FSMG) method mentioned in \cite{van2023anti}, we solve this optimization problem to craft $x'$ as the watermarked image.

\subsection{Activation Token Selection}
\label{sec:tokenSelect}
As previously discussed, the watermark detection phase of our framework requires an activation token to prompt the model to generate images with embedded watermark features. This requires attaching a specific token to the text components associated with the watermarked images in the dataset.
This token functions as an activation trigger, akin to a backdoor embedded in the diffusion model. It is crucial to strike the balance between the stealthiness and effectiveness of these tokens. We present two methods for selecting activation tokens. These methods are practical implementations of the Token-Watermark Alignment strategy and the Watermark-Adding Alignment strategy, respectively, within the proposed framework.

\myparagraph{New Token Construction} An intuitive approach for selecting an activation token is to create a unique word, such as ``lgl''. This word is injected into the prompt text of an image that previously did not contain it. For example, the prompt ``interior of a tavern with violin player'' would be modified to ``lgl, interior of a tavern with violin player'' after injecting the activation token.
Once the diffusion model is fine-tuned using this modified dataset, the token ``lgl'' can effectively trigger the model to generate images with the intended watermark features. While this specifically crafted token can be effective when chosen properly, its conspicuous nature might affect its stealthiness, making it potentially easier for unauthorized users to detect and eliminate.

\myparagraph{Pre-existing Token Utilization} Considering the importance of maintaining high stealthiness in our method, opting to use pre-existing tokens as activation tokens could enhance the method's stealth. In our implementation, we assess the frequency of word token usage across all prompts to identify a word that appears within an appropriate range of total text entries. This approach enables effective control over the proportion of activation token injection, providing a guarantee of both effectiveness and harmlessness. Once we select a suitable word token, we embed the chosen watermark into images associated with this word token in the dataset. This strategy avoids introducing new tokens, thus preserving the original text's stealthiness and preventing noticeable changes that could potentially draw the attention of adversaries.

\subsection{Extension for Large-Scale Authorization}
\label{sec:LargePub}
If a large number of organizations are seeking access to the dataset, it is necessary to design unique activation tokens for different data users to effectively distinguish them. We explore two valid token selection methods in Section~\ref{sec:tokenSelect}. For each method, we need to assess whether the token design approach can handle a large volume of requests.

For the new token construction method, it is easy to achieve the requirement for serving a large number of data users. We can create as many tokens as needed, such as ``lgl'', ``pqp'', ``zxy'', and so on.

For the pre-existing token utilization method, if we rely solely on a single word token based on word frequency to serve as the activation token, the number of unique releases that can be provided is limited. A possible solution is to use the combination of multiple words as an activation token. We can select some single prompt words that meet the conditions and use all corresponding images as those to be watermarked. During the detection phase, we can try these selected single-word tokens in sequence. Formally, let $V$ be the set of all the single-word tokens. For user $i$, the data owner select a collection of single prompt words $T_i = \{\langle \mathsf{token}_1 \rangle,
\langle\mathsf{token}_2 \rangle, ..., \langle \mathsf{token}_{n_i} \rangle\}$, where $\langle \mathsf{token}_j \rangle \in V$. Let $X_i$ be the set of images that will be watermarked. For image $x \in X_i$ and the corresponding text $t$, $x$ belong to $X_i$ if and only if there exists a token in $T_i$, that appears in $t$.  After fine-tuning with the watermarked dataset, we obtain model $M$. The model $M$ is considered to belong to user $i$ if and only if the following condition is satisfied:
$\forall v \in T_i, P(v, M)$ and $\forall v \in V- T_i, \neg P(v, M)$,
where $P(v, M)$ represent that a token $v$ successfully triggers model $M$. The processes of distributing data to users and tracing the leaker are described in detail in Algorithm~\ref{algorithm-1} and Algorithm~\ref{algorithm-2}, respectively.

\begin{table}[]
\centering
\footnotesize
\caption{Results on effectiveness for TWA and WAA methods. }
\label{tab:effectiveness}
\begin{tabular}{cc|cc|cc}
\hline
\multicolumn{2}{c|}{Dataset}                                                      & \multicolumn{2}{c|}{WikiArt} & \multicolumn{2}{c}{COCO} \\ \hline
\multicolumn{2}{c|}{Fine-Tuning Method}                                           & Standard       & LoRA        & Standard     & LoRA      \\ \hline\hline
\multicolumn{1}{c|}{\multirow{3}{*}{\makecell[c]{TWA \\(0.016)}}} & TWA-G   & 100\%          & 85.3\%      & 99.3\%       & 61.5\%    \\
\multicolumn{1}{c|}{}                                                   & TWA-DWT & 100\%          & 82.6\%      & 100\%        & 64.4\%    \\
\multicolumn{1}{c|}{}                                                   & TWA-Adv & 100\%          & 100\%       & 100\%        & 96.5\%    \\ \hline
\multicolumn{1}{c|}{\multirow{3}{*}{\makecell[c]{TWA \\(0.256)}}} & TWA-G   & 100\%          & 87.3\%      & 99.6\%       & 58.7\%    \\
\multicolumn{1}{c|}{}                                                   & TWA-DWT & 100\%          & 98.2\%      & 100\%        & 87.4\%    \\
\multicolumn{1}{c|}{}                                                   & TWA-Adv & 100\%          & 100\%       & 100\%        & 100\%     \\ \hline
\multicolumn{1}{c|}{\multirow{3}{*}{WAA}}        & WAA-G   & 98.9\%         & 58.1\%      & 89.5\%       & 56.9\%    \\
\multicolumn{1}{c|}{}                                                   & WAA-DWT & 99.8\%         & 58.0\%      & 96.9\%       & 64.7\%    \\
\multicolumn{1}{c|}{}                                                   & WAA-Adv & 100\%          & 96.1\%      & 100\%        & 91.3\%    \\ \hline
\end{tabular}
\end{table}

\section{Evaluation}
We conduct extensive experiments on image generation tasks by fine-tuning Stable Diffusion models to demonstrate the efficiency of our framework. We begin by describing our experimental setup in Section~\ref{sec:exp:setup}. Next, we evaluate the effectiveness and harmlessness of our approach in Sections~\ref{sec:watermarkingeffectiveness} and \ref{sec:exp:quality}, respectively. In addition, we evaluate the transferability of the proposed framework in Section~\ref{sec:exp:trans} and discuss the results of large-scale authorization in Section~\ref{sec:traceability}, including the robustness and traceability of the proposed method.
Some examples demonstrating the stealthiness of the proposed method have been shown in Figure~\ref{fig:Example-1}, with additional examples provided in the appendices. Moreover, our method requires modifying only about 2\% of the data to achieve high detection accuracy. The hyperparameters are provided in Appendix~\ref{sec:hyperparameters}.

\subsection{Experimental Setup}
\label{sec:exp:setup}
Our method is implemented by using Python 3.10.0 and PyTorch 2.2.2. All experiments are conducted on a server with Ubuntu 22.04 system, equipped with two Nvidia RTX 6000 GPU cards.

\myparagraph{Model} We utilize Stable Diffusion v2-1~\cite{Rombach_2022_CVPR}\footnote{https://huggingface.co/stabilityai/stable-diffusion-2-1} as our text-to-image diffusion model in the experiments. We attempt to make use of two kinds of fine-turning methods: standard full parameter fine-tuning and LoRA~\cite{hu2021lora}. The former is a powerful method for fine-tuning the model by updating all the parameters, while the latter offers a lightweight option.  We use pretrained ResNet34 model~\cite{he2016deep} for binary classification of the images in watermarking detection.

\myparagraph{Datasets} We conduct our experiments on two datasets: the WikiArt dataset~\cite{saleh2015large} and the COCO dataset~\cite{lin2014microsoft}. The WikiArt dataset consists of 52,757 paintings by 195 artists, spanning 27 genres, each accompanied by a caption detailing its content. The COCO dataset includes over 330,000 images, with 220,000 annotated, encompassing 1.5 million objects across 80 object categories. For the task of protecting data with specific artwork styles, we randomly select 1,000 samples from the  ``Baroque'' genre in the WikiArt dataset. For the task of protecting data with specific subjects, we randomly choose 1,000 samples from the ``Cat'' category in the COCO dataset. Figure~\ref{fig:stealthinessfig} in the appendices shows examples from the WikiArt and COCO datasets before and after watermarking.

\myparagraph{Methods} We consider three types of watermarking schemes described in Section~\ref{sec:watermarkingSelect} under two strategies: the Token-Watermark Alignment strategy and the Watermark-Adding Alignment strategy and their corresponding concrete implementations described in Section~\ref{sec:tokenSelect}. Specifically, we evaluate the following six methods in two categories in our experiments. The first category includes the Token-Watermark Alignment strategy, implemented by the new token construction method with Gaussian noise watermarking (\textbf{TWA-G}), DWT watermarking (\textbf{TWA-DWT}), and adversarial example watermarking (\textbf{TWA-Adv}). The second category includes the Watermark-Adding Alignment strategy, implemented by the pre-existing token utilization method with Gaussian noise watermarking (\textbf{WAA-G}), DWT watermarking (\textbf{WAA-DWT}), and adversarial example watermarking (\textbf{WAA-Adv}).

In the first category of methods, we can change a small amount of text prompts in the dataset. When a data user requests the dataset, we select the token ``lgl'' as the special token and put it at the beginning of the corresponding text prompts of the specific images.
In the second category of methods, we can select a specific token from the pre-existing tokens without modifying any text prompts, ensuring that the frequency of this token remains low. Considering that the capability of the LoRA method is less than that of the standard fine-tuning method when selecting tokens, for the LoRA method, we slightly increased the token frequency. This adjustment helps us effectively control the proportion of token injection. The details of the token selection are presented in Table~\ref{table:BackdoorWikiart} and Table~\ref{table:BackdoorCOCO}  in the Appendix\ref{sec:AppendTokenSelect}.

\begin{figure}[!t]
    \centering
    \begin{subfigure}[b]{0.19\textwidth}
       \includegraphics[width=\textwidth]{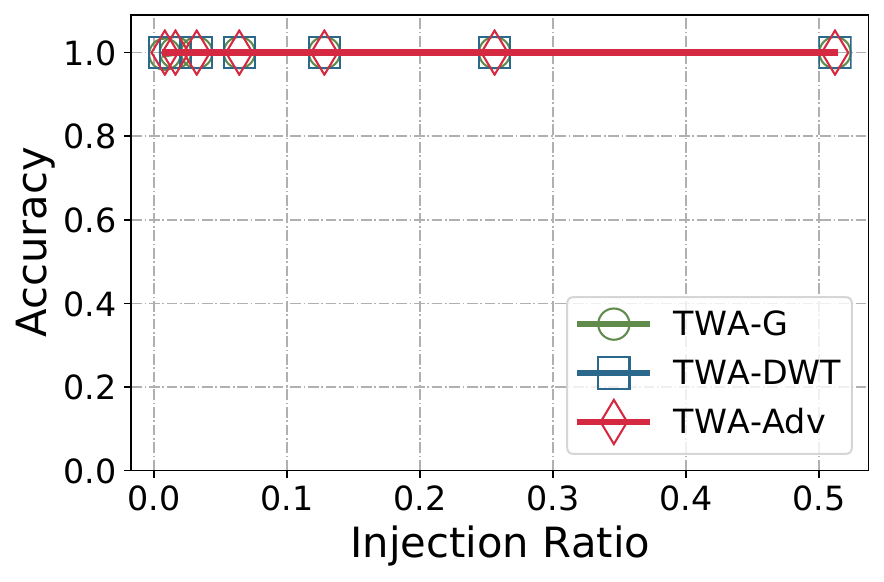}
        \caption{WikiArt, Standard}
        \label{fig:clstask123ST}
    \end{subfigure}
    \begin{subfigure}[b]{0.19\textwidth}
        \includegraphics[width=\textwidth]{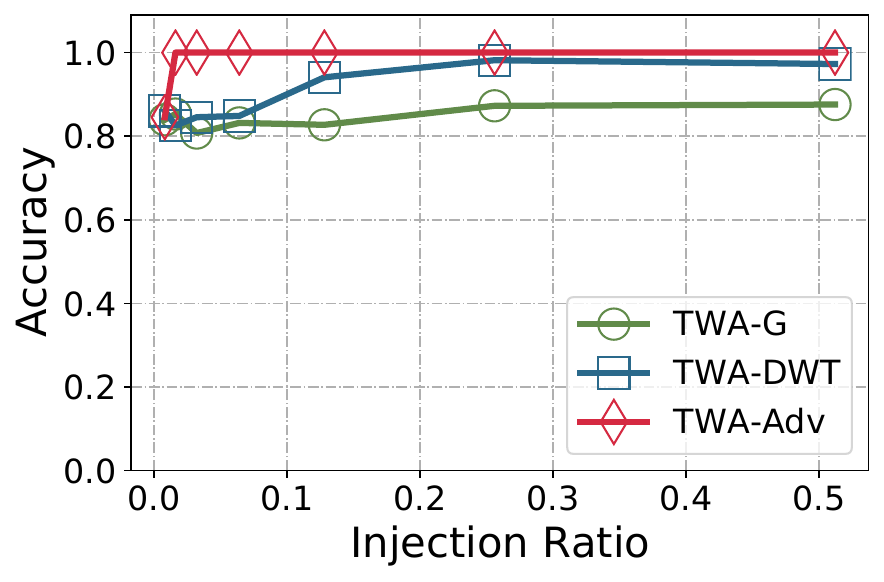}
        \caption{WikiArt, LoRA}
         \label{fig:clstask123LORA}
    \end{subfigure}

    \begin{subfigure}[b]{0.19\textwidth}
        \includegraphics[width=\textwidth]{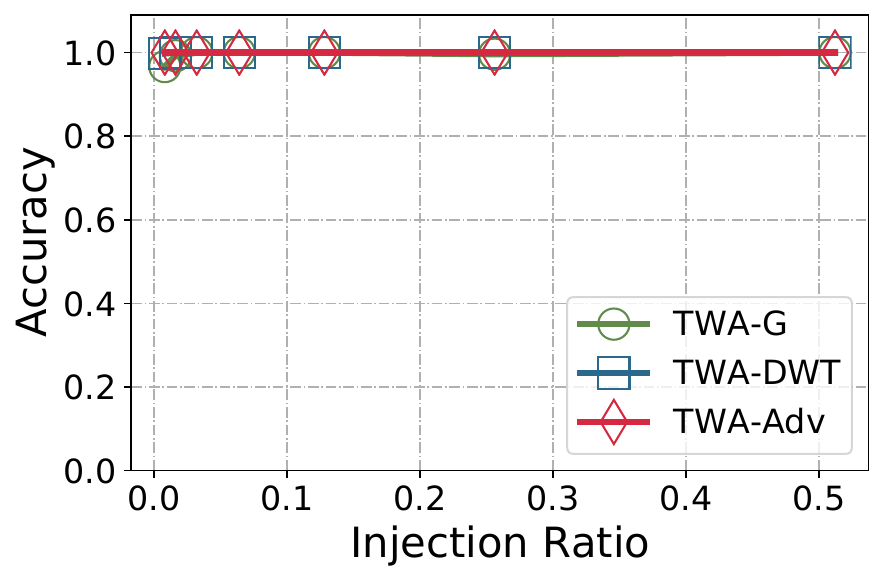}
        \caption{COCO, Standard}
         \label{fig:clstask123LORA}
    \end{subfigure}
    \begin{subfigure}[b]{0.19\textwidth}
        \includegraphics[width=\textwidth]{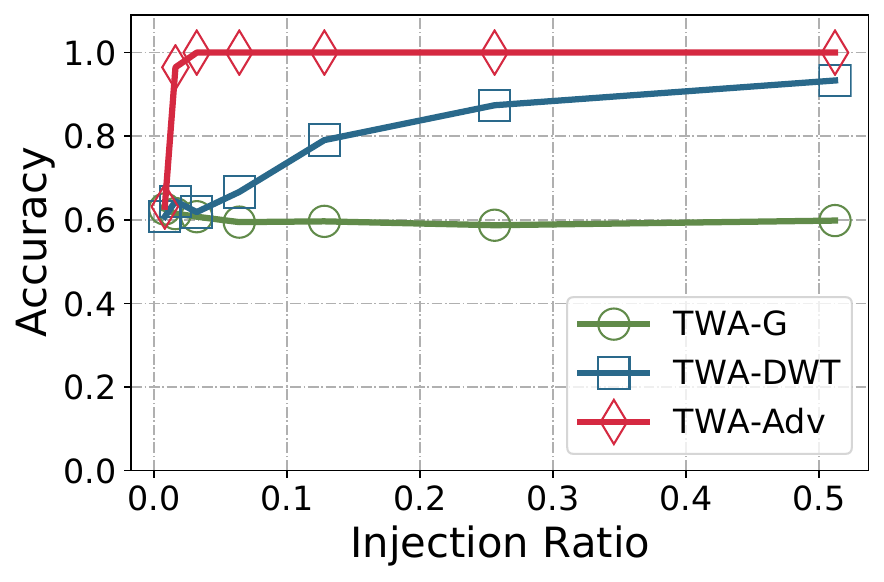}
        \caption{COCO, LoRA}
         \label{fig:clstask123LORA}
    \end{subfigure}
    \caption{TWA: detection accuracy w/ various injection ratios}
    \label{fig:clstask123}
\end{figure}

\myparagraph{Metrics} We measure the effectiveness of the proposed framework by the accuracy of watermark detection. An effective method is characterized by the ability to differentiate between images created by a target text-to-image model using either prompts with a specific activation token or regular prompts. Additionally, the quality of the text-to-image synthesis is measured using the Fr\'echet Inception Distance (FID)~\cite{heusel2017gans}. This metric compares images generated by the model trained on the original dataset with those generated by the model trained on the protected dataset by our framework. Similar FID scores between these two sets of generated images suggest that the protection has minimal impact on image quality.

\subsection{Watermarking Effectiveness}
\label{sec:watermarkingeffectiveness}
\myparagraph{TWA Methods} Specifically, for the Token-Watermark Alignment methods, we prepend a specially constructed token (``lgl'') to the beginning of the text prompts of the first $8$, $16$, $32$, $64$, $128$, $256$, and $512$ images, respectively. For TWA-G, TWA-DWT, and TWA-Adv, we add the corresponding watermarks to the selected images. We have trained seven models for each method on each dataset with respect to different levels of injection ratio. We use the substitute model to generate the datasets for the detector. We selected $1,440$ images for the training set and $540$ images for the test set. Specifically, the prompts used to generate images are as follows:
\begin{description}[leftmargin=1.0cm,noitemsep,topsep=0pt,parsep=0pt,partopsep=0pt]
    \item $\lbrack$ >\_ $\rbrack$ Prompt-1:
    \begin{description}[leftmargin=1.0cm,noitemsep,topsep=0pt,parsep=0pt,partopsep=0pt]
        \item ``\textit{A painting in the style of Baroque}'' (WikiArt)
        \item ``\textit{A photo of a cat}'' (COCO)
    \end{description}
    \item $\lbrack$ >\_ $\rbrack$ Prompt-2:
    \begin{description}[leftmargin=1.0cm,noitemsep,topsep=0pt,parsep=0pt,partopsep=0pt]
        \item ``\textit{lgl, A painting in the style of Baroque}'' (WikiArt)
        \item ``\textit{lgl, A photo of a cat}'' (COCO)
    \end{description}
\end{description}

\begin{figure}[!t]
    \centering
    \begin{subfigure}[b]{0.19\textwidth}
       \includegraphics[width=\textwidth]{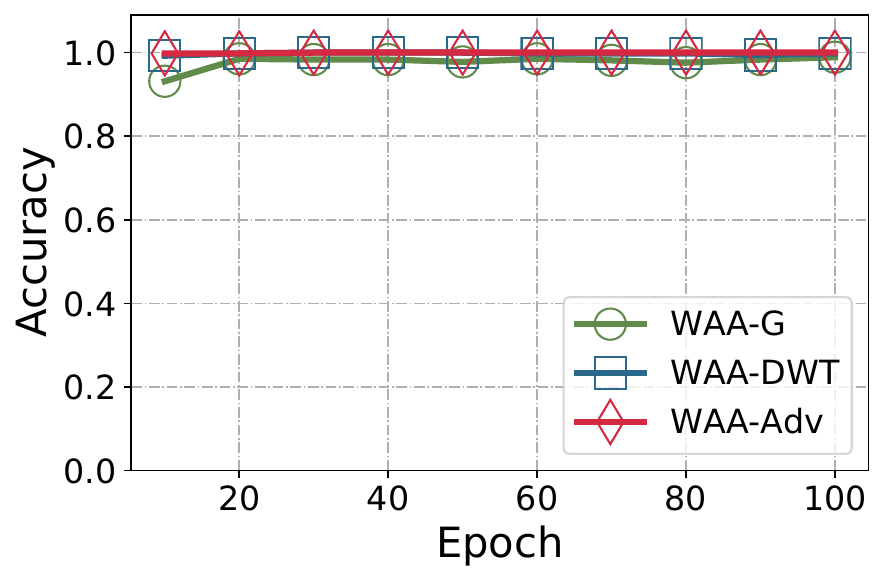}
        \caption{WikiArt, Standard}
        \label{fig:clstask456ST}
    \end{subfigure}
    \begin{subfigure}[b]{0.19\textwidth}
        \includegraphics[width=\textwidth]{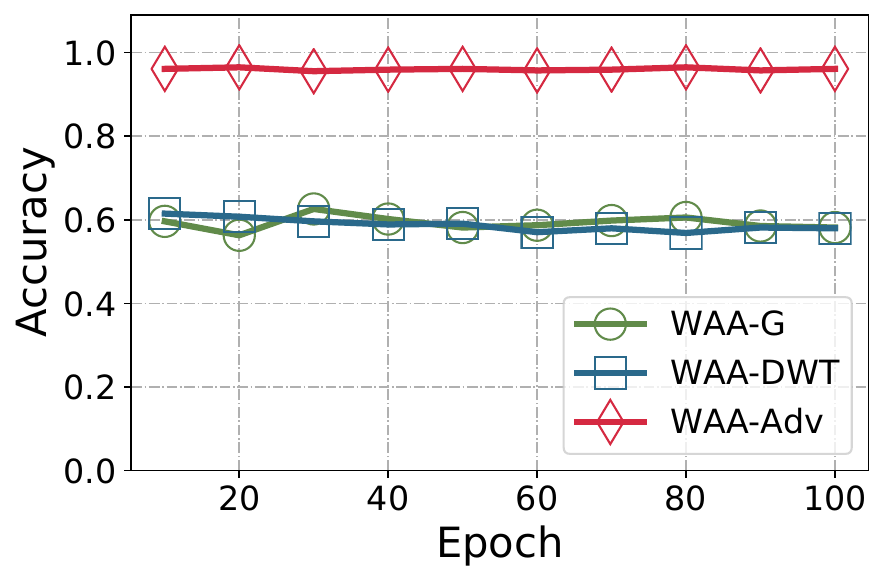}
        \caption{WikiArt, LoRA}
         \label{fig:clstask456LORA}
    \end{subfigure}

    \begin{subfigure}[b]{0.19\textwidth}
        \includegraphics[width=\textwidth]{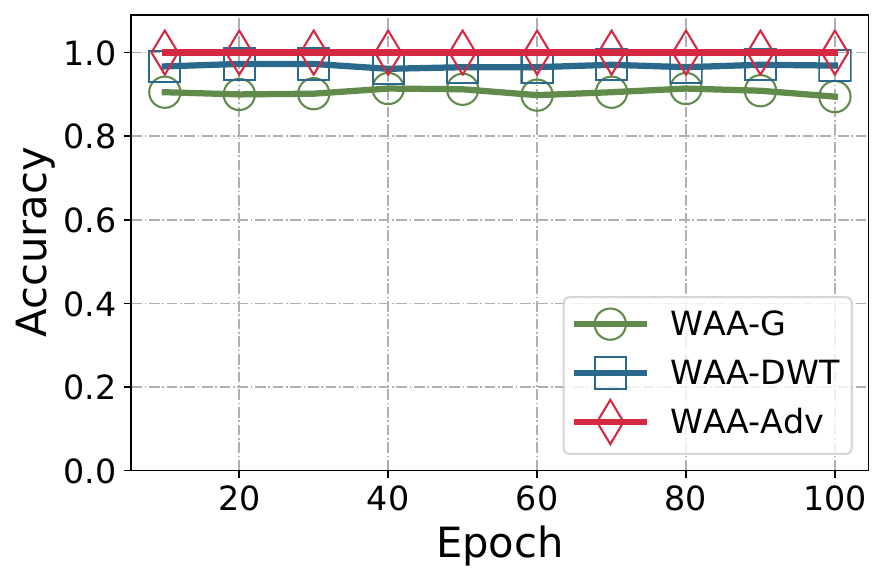}
        \caption{COCO, Standard}
         \label{fig:clstask456LORA}
    \end{subfigure}
    \begin{subfigure}[b]{0.19\textwidth}
        \includegraphics[width=\textwidth]{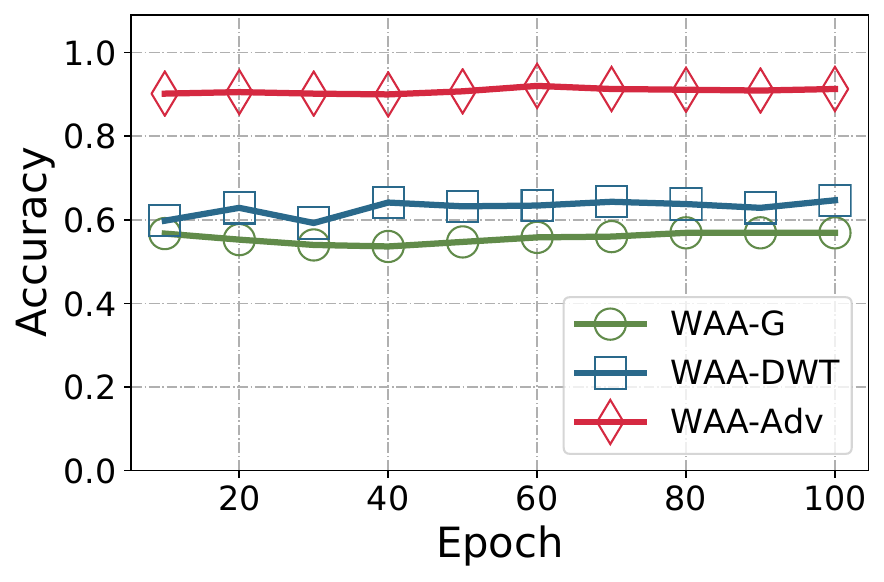}
        \caption{COCO, LoRA}
         \label{fig:clstask456LORA}
    \end{subfigure}
    \caption{WAA: detection accuracy w/ various training epochs}
    \label{fig:task456ACC}
\end{figure}

Some of the generated images are presented in the Appendix~\ref{sec:TWAImages}. The results are shown in Table~\ref{tab:effectiveness}, where we evaluate the TWA methods with a low injection ratio of 0.016 and a high injection ratio of 0.256.
For different settings, in general, TWA-Adv performs the best, followed by TWA-DWT, with TWA-G performing the least satisfactory. For the standard fine-tuning method, all three TWA methods can achieve nearly 100\% accuracy on both datasets. For the LoRA fine-tuning method, the detection accuracy will be significantly affected by the choice of protection method, and we can find that TWA-Adv can still achieve an accuracy higher than 95\% for all the settings. We also evaluate the TWA methods by changing the injection ratio, and the results are shown in Figure~\ref{fig:clstask123}. We can observe that our methods, in particular the TWA-Adv method, can reach a high detection accuracy with a very small injection ratio, which implies a high level of stealthiness of the proposed methods.

\myparagraph{WAA Methods}
As expected, for a target model, only the images generated using the specific activation token will trigger the watermark, differentiating it from the other prompts. We train 10 text-to-image models by fine-tuning the pretrained Stable Diffusion for each method on each dataset using different selected activation tokens, as detailed in Table~\ref{table:BackdoorWikiart} and Table~\ref{table:BackdoorCOCO} in the Appendix~\ref{sec:AppendTokenSelect}. For each model, we use these 10 activation tokens as prompts to generate 100 images each, resulting in a total of 1,000 images generated per model for each method on each dataset. As mentioned in Section~\ref{discriminator}, we trained only one unified detector for all data users in each experimental setting (each method with each fine-tuning method on each dataset).

The results are detailed in Table~\ref{tab:effectiveness} and Figure~\ref{fig:task456ACC}. For the standard training method, the WAA-Adv method shows extremely high detection accuracy, achieving 100\% detection accuracy on both datasets. On the two datasets, the WAA-G and WAA-DWT methods achieved classification accuracies of 98.9\%/89.5\% and 99.8\%/96.9\%, respectively. For the LoRA method, although the WAA-G and WAA-DWT methods do not achieve satisfactory accuracy, the WAA-Adv method can still achieve high detection accuracy, reaching 96.1\% and 91.3\% on the two datasets, respectively. The protection is less effective under LoRA fine-tuning because this method modifies the model in a more restricted way compared to standard fine-tuning, which optimizes parameters more comprehensively. This constrained adjustment leads to outputs that are less detailed and nuanced. Consequently, watermarks in images generated by models fine-tuned with LoRA are less recognizable. We also include the results of the convergence of training detectors for the WAA methods, which are shown in Figure~\ref{fig:task456ACC}. We can observe that the detector can be trained with very few epochs, achieving a fast training speed.
For the WAA methods, accuracy is somewhat lower compared to the TWA methods, which involve injecting new tokens into the datasets. However, WAA methods maintain higher stealthiness as they do not modify the text, and still achieve satisfactorily high classification accuracy.
Some of the generated images are presented in the appendices.


\begin{figure}[!t]
    \centering
    \begin{subfigure}[b]{0.19\textwidth}
       \includegraphics[width=\textwidth]{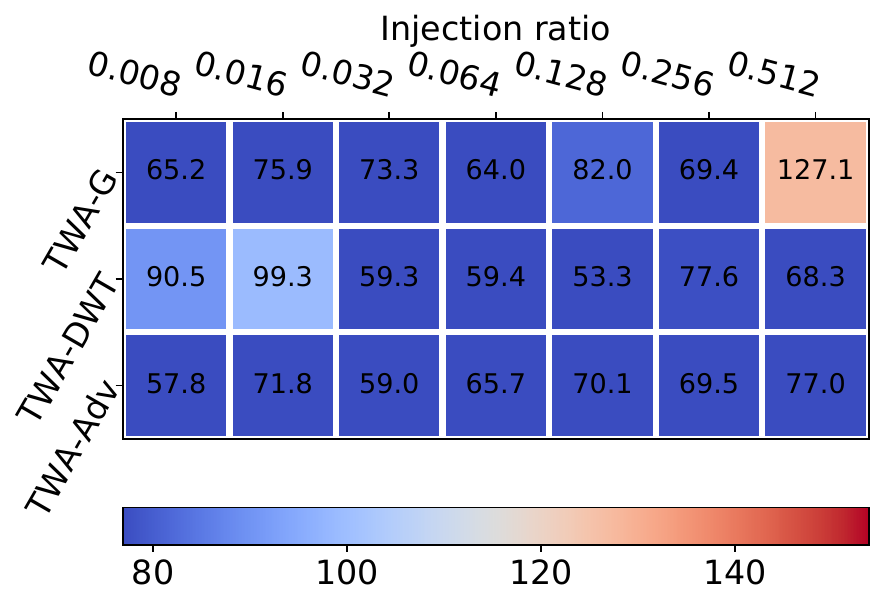}
        \caption{WikiArt, Standard}
        \label{fig:fid-clstask123ST}
    \end{subfigure}
    \begin{subfigure}[b]{0.19\textwidth}
        \includegraphics[width=\textwidth]{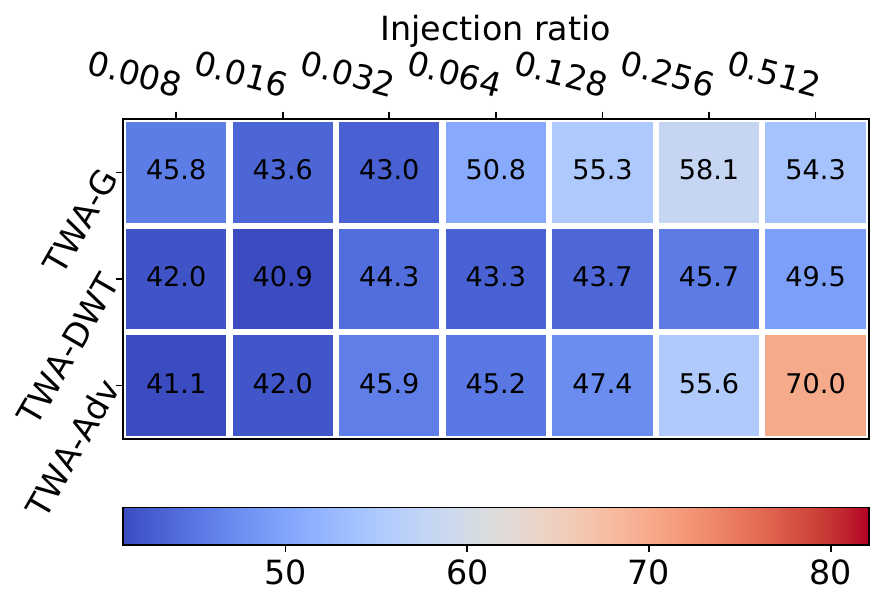}
        \caption{WikiArt, LoRA}
         \label{fig:fid-clstask123LORA}
    \end{subfigure}

    \begin{subfigure}[b]{0.19\textwidth}
        \includegraphics[width=\textwidth]{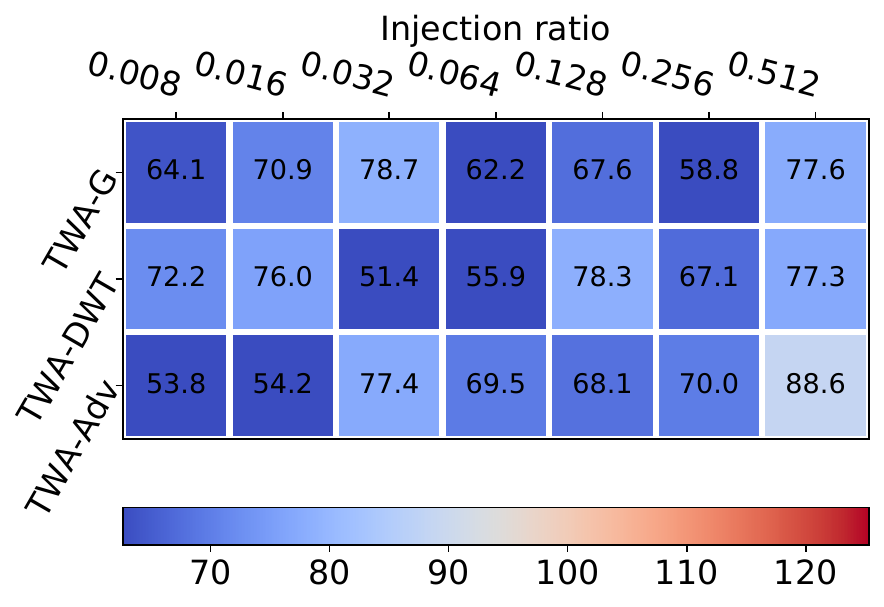}
        \caption{COCO, Standard}
         \label{fig:fid-clstask123LORA}
    \end{subfigure}
    \begin{subfigure}[b]{0.19\textwidth}
        \includegraphics[width=\textwidth]{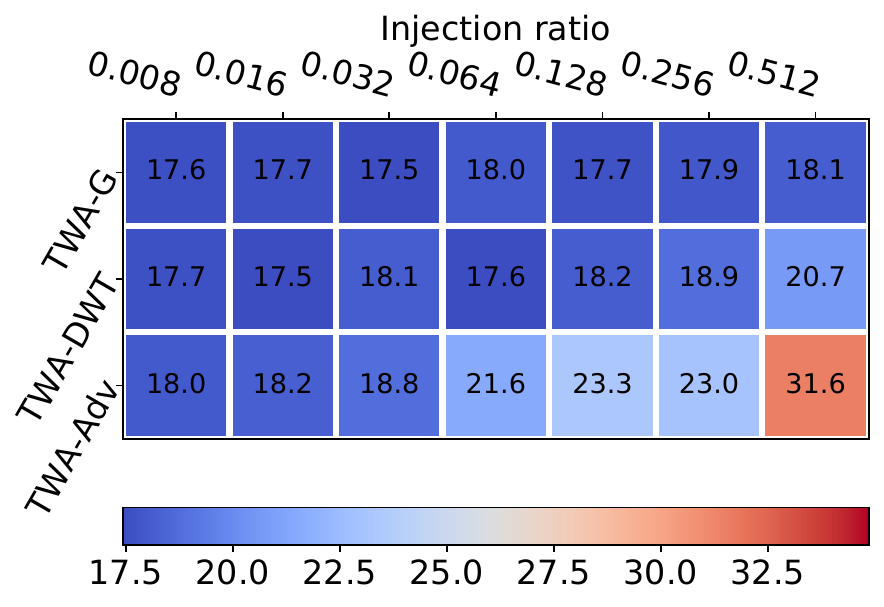}
        \caption{COCO, LoRA}
         \label{fig:fid-clstask123LORA}
    \end{subfigure}
    \caption{FID results for evaluating the quality of the generated image: TWA methods.}
    \label{fig:FIDGCO-TWA}
\end{figure}

\subsection{Image Synthesis Quality}
\label{sec:exp:quality}
While ensuring the effectiveness of the generated watermark, it is crucial to maintain that the watermarked dataset can still produce high-quality images under normal usage. Similarly, we choose the prompts ``\textit{A painting in the style of Baroque}'' and ``\textit{A photo of a cat}'' for WikiArt and COCO datasets, respectively, to evaluate the synthesis quality. Some of the generated images are presented in the Appendix~\ref{sec:appendImageQuantity}.

For each dataset and each fine-tuning method, we have trained a text-to-image synthesis model by using original images without protection and generated 2,000 benchmark images. We have divided the images into two sets each with 1,000 images. We calculate the FID between these two sets of images to establish a baseline of the FID scores. For the standard fine-tuning method and the LoRA method on the WikiArt dataset, the FID scores are $76.89$ and $41.05$, respectively. For the standard fine-tuning method and the LoRA method on the COCO dataset, the FID scores are $62.65$ and $17.43$, respectively.

\begin{figure}[!t]
    \centering
    \begin{subfigure}[b]{0.19\textwidth}
       \includegraphics[width=\textwidth]{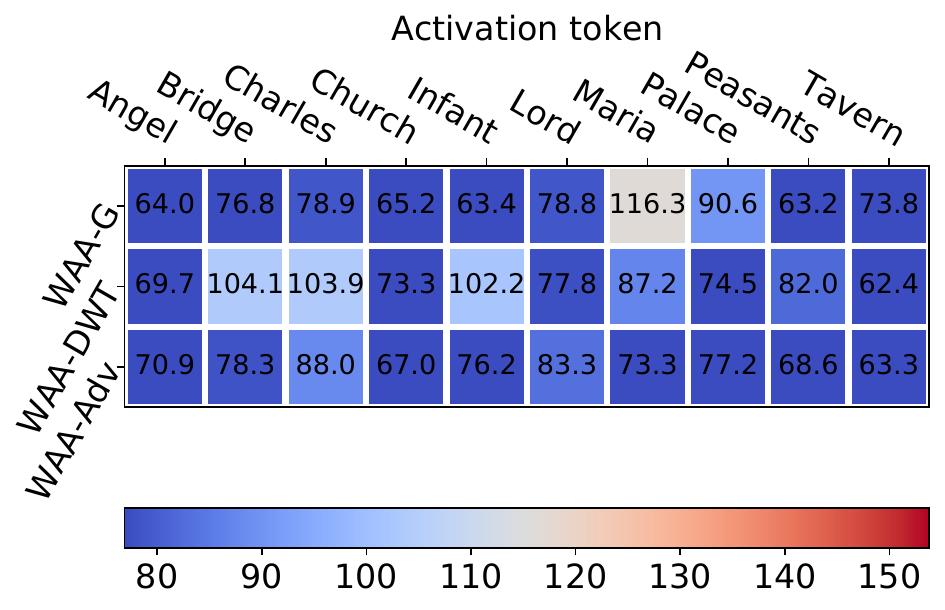}
        \caption{WikiArt, Standard}
        \label{fig:fid-clstask456ST}
    \end{subfigure}
    \begin{subfigure}[b]{0.19\textwidth}
        \includegraphics[width=\textwidth]{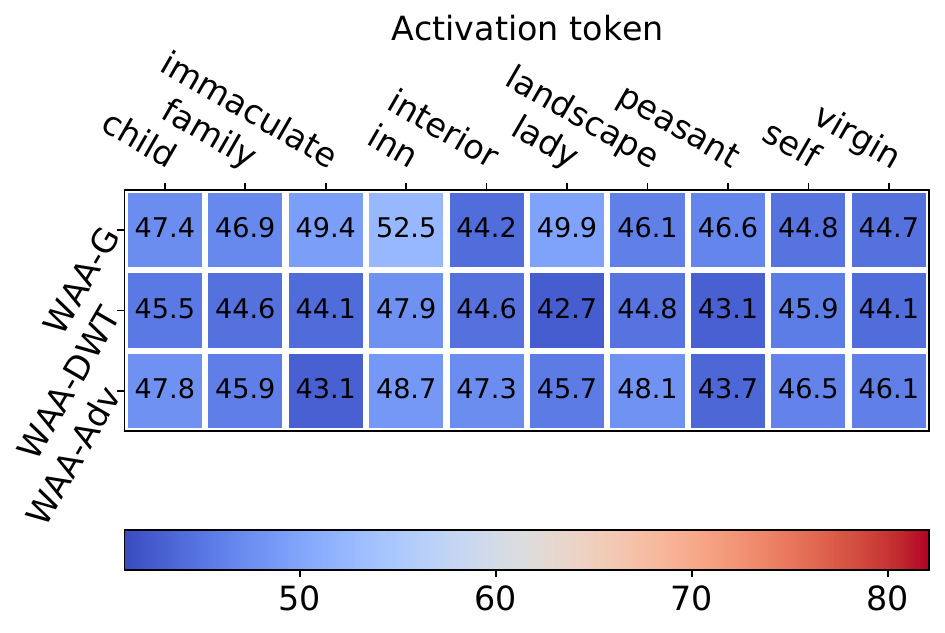}
        \caption{WikiArt, LoRA}
         \label{fig:fid-clstask456LORA}
    \end{subfigure}

    \begin{subfigure}[b]{0.19\textwidth}
        \includegraphics[width=\textwidth]{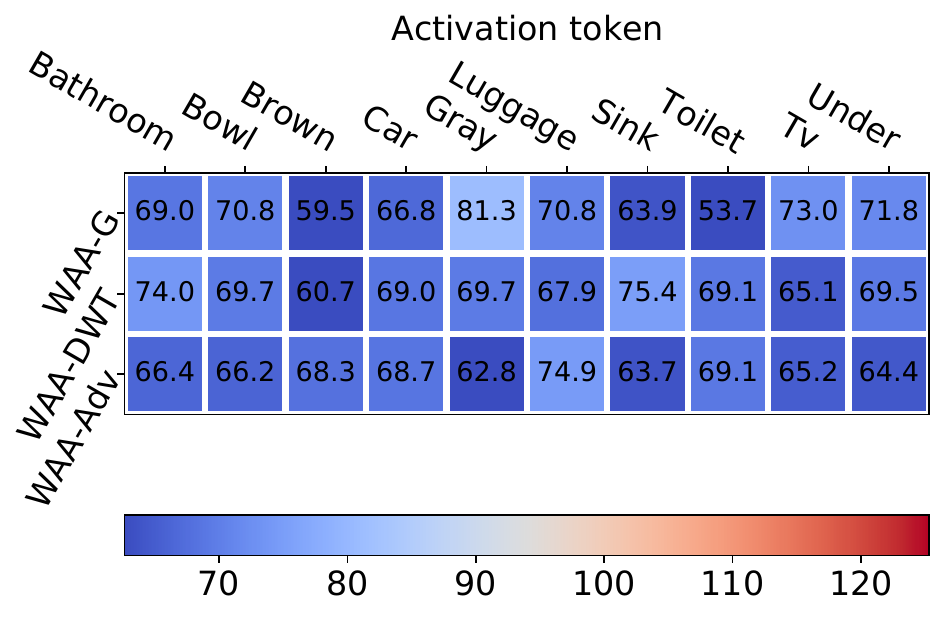}
        \caption{COCO, Standard}
         \label{fig:fid-clstask456LORA}
    \end{subfigure}
    \begin{subfigure}[b]{0.19\textwidth}
        \includegraphics[width=\textwidth]{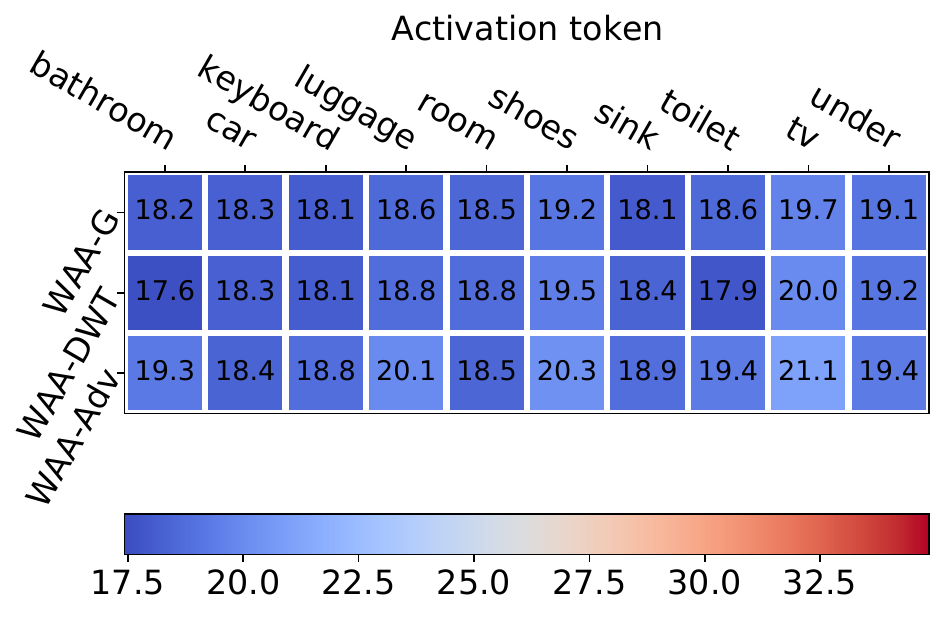}
        \caption{COCO, LoRA}
         \label{fig:fid-clstask456LORA}
    \end{subfigure}
    \caption{FID results for evaluating the quality of the generated image: WAA methods.}
    \label{fig:FIDGCO-WAA}
\end{figure}

We have trained a text-to-image model on the watermarked dataset for each method and each dataset. We use the same prompt to generate 1,000 images for each model.
Subsequently, we have calculated the FID values between the images generated under the setting of each method and each dataset and the corresponding set of benchmark images. For TWA methods, we have evaluated the quality of generated images with all the injection ratios mentioned above. For WAA methods, we have evaluated the model trained on each activation token selected. The results are detailed in Figure~\ref{fig:FIDGCO-TWA} and Figure~\ref{fig:FIDGCO-WAA}. From the experimental results, the FID of the images generated under the normal prompt by models trained on the protected datasets is very close to the benchmark images, indicating that the protection method has almost no negative effect on the text-to-image synthesis task.

\begin{table}[h!]
\centering
\caption{Transferability across different datasets. The result ``Acc. Benchmark'' refers to the highest achievable accuracy of a single dataset as detailed in Section \ref{sec:watermarkingeffectiveness}, serving as a benchmark for the detection performance.}
\footnotesize
\begin{tabular}{c|c|c|c}
\hline
Setting & TPR & Acc. & Acc. Benchmark\\
\hline\hline
STD $\rightarrow$ LoRA, WikiArt & 81.5\% & 88.0\% & 96.1\% \\
\hline
STD $\rightarrow$ LoRA, COCO & 85.7\% & 80.4\% & 91.3\% \\
\hline
STD, WikiArt $\rightarrow$ COCO & 98.2\% & 99.0\% & 100\% \\
\hline
STD, COCO $\rightarrow$ WikiArt & 100\% & 93.3\% & 100\% \\
\hline
\end{tabular}
\label{table:classifierPerformance}
\end{table}

\begin{table}[h!]
\centering
\caption{Transferability performance.}
\footnotesize
\begin{tabular}{c|c|c|c|c}
\hline
\makecell[c]{Model \\ Dataset} & \makecell[c]{Testing \\ Dataset} & TPR & Acc. & \makecell[c]{Acc. \\Benchmark} \\
\hline\hline
WikiArt & LoRA-TWA-COCO & 100\% & 99.8\% & 100\% \\
\hline
WikiArt & LoRA-WAA-COCO & 88.5\% &  84.1\% & 91.3\% \\
\hline
WikiArt & STD-TWA-COCO & 100\% & 100\% & 100\% \\
\hline
WikiArt & STD-WAA-COCO & 100\% & 99.2\% & 100\% \\
\hline
COCO & LoRA-TWA-WikiArt & 100\% & 100\% & 100\% \\
\hline
COCO & LoRA-WAA-WikiArt & 97.8\% & 90.6\% & 96.1\% \\
\hline
COCO & STD-TWA-WikiArt & 100\% &100\% & 100\% \\
\hline
COCO & STD-WAA-WikiArt & 84.4\% & 92.2\% & 100\% \\
\hline
\end{tabular}
\label{table:modelPerformance}
\end{table}

\subsection{Transferability}
\label{sec:exp:trans}
In this section, we examine the transferability of the proposed methods. Since data users may employ various fine-tuning techniques and the watermarking detector may be trained on different datasets, ensuring the transferability of the proposed methods is crucial. Achieving high detection accuracy across different training methodologies and datasets can be challenging due to the high variability of watermarking features under different conditions. Our objective is to demonstrate the robust transferability of our methods.

We assume that the data owner can access suspected models and has the option to use either the TWA or WAA methods to determine the activation token. Our primary aim is to evaluate the transferability of the proposed methods across different fine-tuning methods and datasets. We utilize the detector with the WAA method and full parameter fine-tuning (STD) technique on the COCO and WikiArt datasets. Transferability is assessed under various settings, with the results presented in Table~\ref{table:classifierPerformance}. In the table, the first part of each setting indicates the fine-tuning method or the use of the detector across different fine-tuning methods, while the second part specifies the dataset or the use of the detector across different datasets. Our results indicate that the WAA method performs consistently well in terms of both accuracy and true positive rate (TPR) across various fine-tuning methods and datasets, showing the strong transferability of our methods. When compared to the Acc. benchmark, our detector exhibits a maximum accuracy drop of 10.9\% and an average decrease of 5.0\%.

We note that the visibility of adversarial features in images is evident even to the human eye. This has inspired us to train a general binary classifier through a subject dataset. Then we test the transferability of this detector across different datasets. Specifically, we randomly selected 100 images from each of the following categories: STD-WAA-WikiArt, STD-TWA-WikiArt, LoRA-WAA-WikiArt, and LoRA-TWA-WikiArt, classifying them into the adversarial output class. The same number from the normal output class is randomly selected similarly. The classifier has been trained on these 800 images to distinguish between adversarial and normal outputs, applicable to both standard and LoRA fine-tuning methods. Subsequently, we assessed the transferability of this detector on the COCO dataset and conducted similar tests in reverse. The results are presented in Table~\ref{table:modelPerformance}.

The maximum deviation between the benchmark accuracy and the observed accuracy is approximately 7.8\%, with an average difference of about 2.6\%. These results demonstrate that our general detector exhibits robust transferability across various datasets and fine-tuning methods, significantly enhancing the practical deployment of our methods.

\subsection{Large-Scale Authorization}
\label{sec:traceability}
As mentioned in Section~\ref{sec:LargePub}, if a large number of organizations are seeking access to the data, for the TWA methods, we can construct an infinite number of activation tokens to meet our needs. For the WAA methods, we can use the combination of multiple words as an activation token. The combination activation tokens are detailed in Table~\ref{table:BackdoorComb} in the Appendix\ref{sec:AppendTokenSelect}. In this part, we evaluate our method for large-scale authorization.

\begin{figure}[!t]
    \centering
    \begin{subfigure}[b]{0.20\textwidth}
       \includegraphics[width=\textwidth]{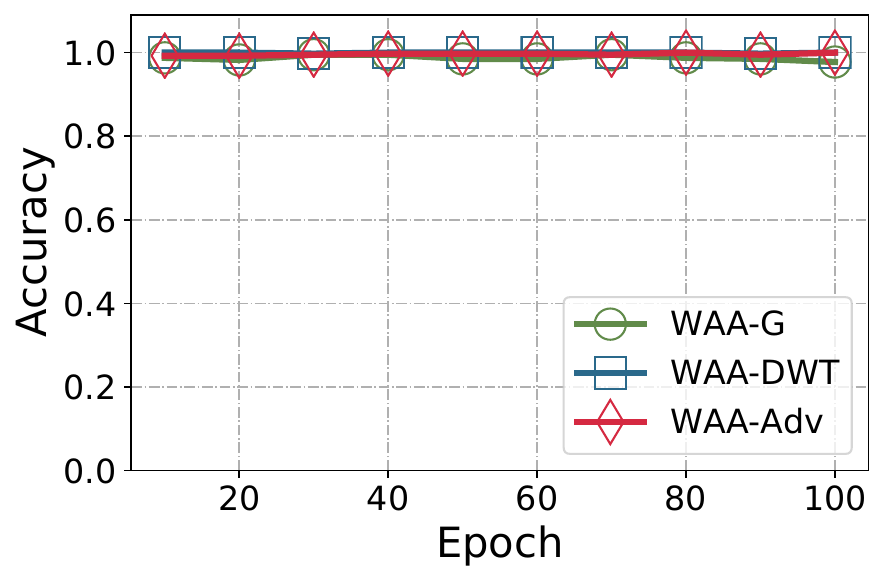}
        \caption{WikiArt, Standard}
    \end{subfigure}
    \begin{subfigure}[b]{0.20\textwidth}
        \includegraphics[width=\textwidth]{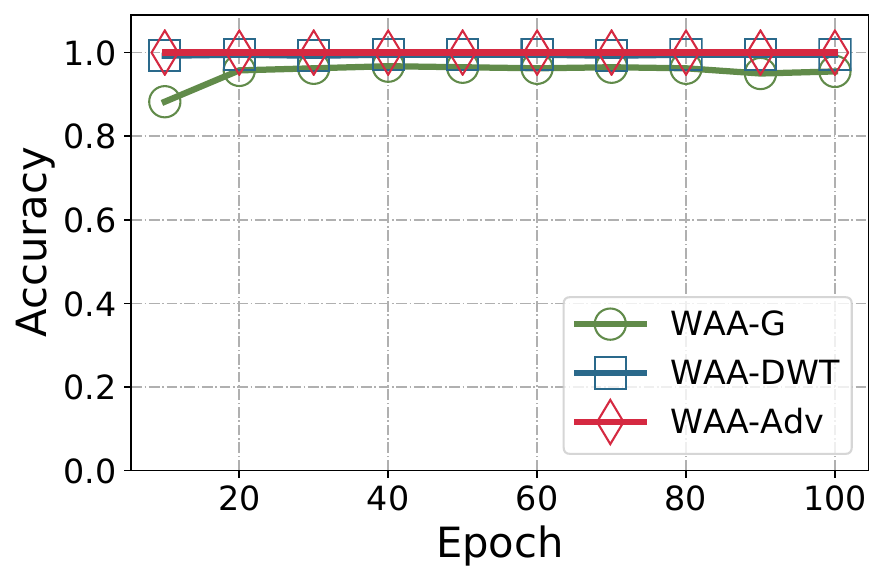}
        \caption{COCO, Standard}
         \label{fig:COCO-comb}
    \end{subfigure}
    \caption{Effectiveness for the large-scale authorization.}
    \label{fig:comb-acc}
\end{figure}

\begin{figure}[tb]
  \centering
\begin{subfigure}[b]{0.22\textwidth}
       \includegraphics[width=\textwidth]{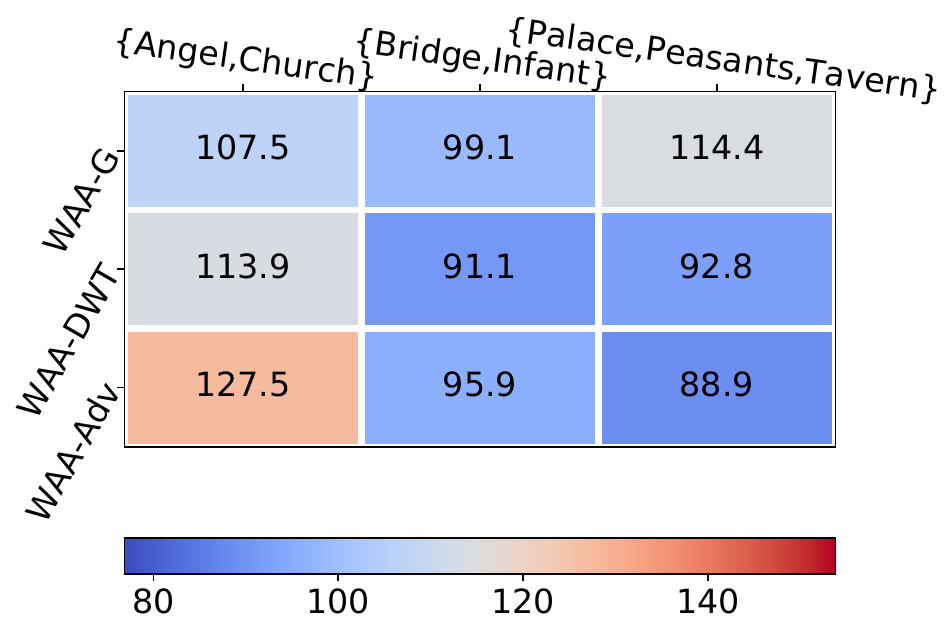}
        \caption{WikiArt, Standard}
    \end{subfigure}
    \begin{subfigure}[b]{0.20\textwidth}
        \includegraphics[width=\textwidth]{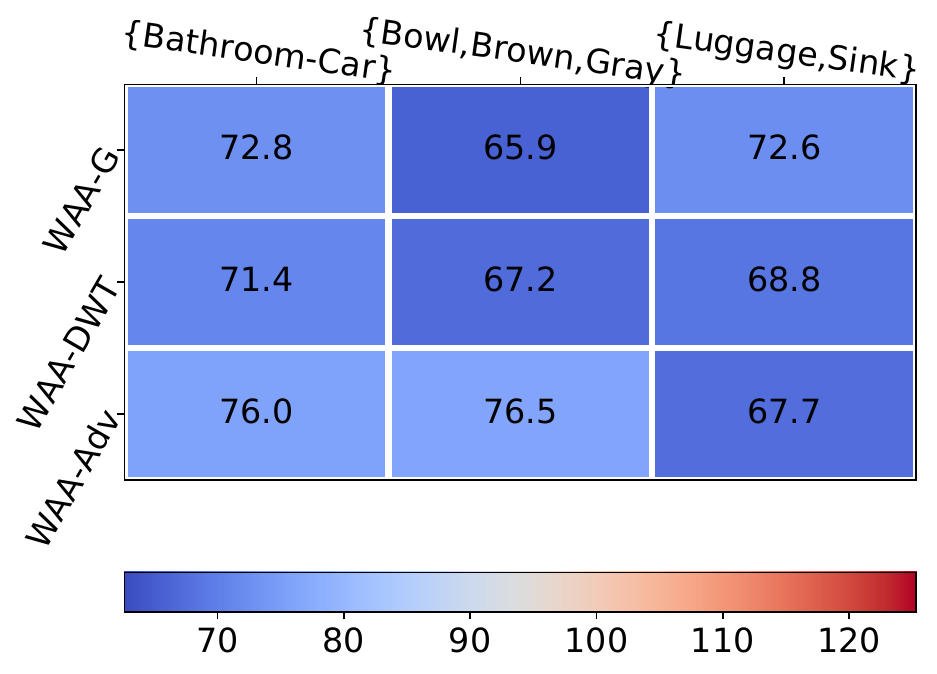}
        \caption{COCO, Standard}
         \label{fig:COCO-comb}
    \end{subfigure}
    \caption{FID for the large-scale authorization.}
    \label{fig:comb-fid}
\end{figure}

\begin{figure*}[!t]
    \centering
    \begin{subfigure}[b]{0.19\textwidth}
       \includegraphics[width=\textwidth]{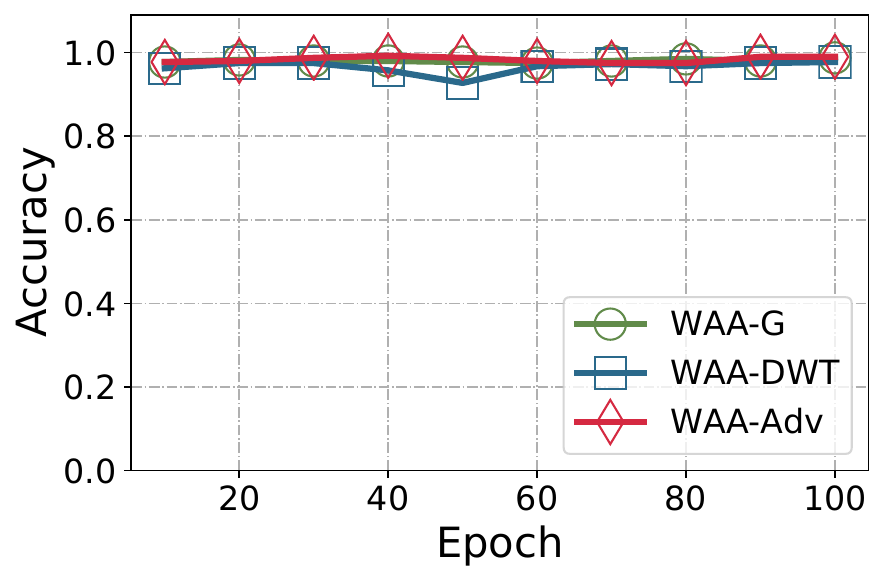}
        \caption{JPEG Compression}
        \label{fig:compST-1}
    \end{subfigure}
    \begin{subfigure}[b]{0.19\textwidth}
        \includegraphics[width=\textwidth]{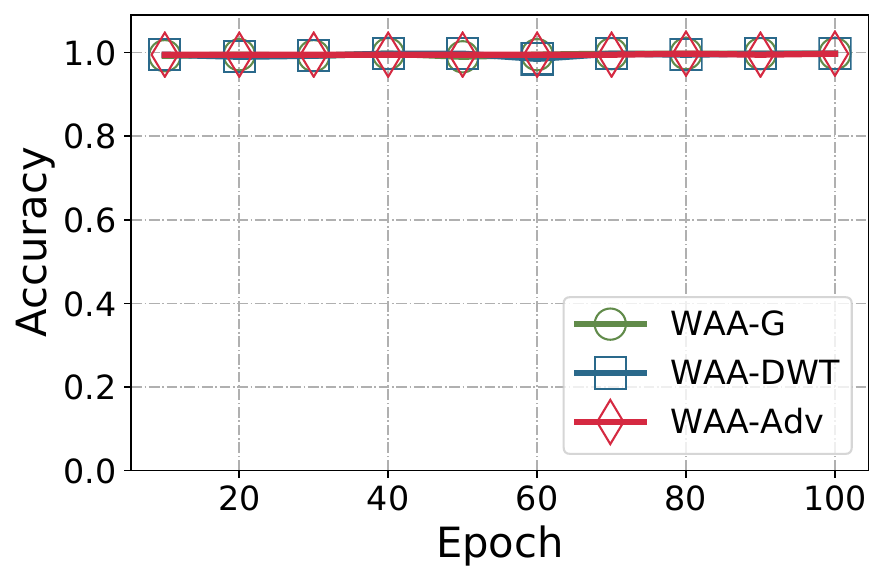}
        \caption{Sharpness}
         \label{fig:sharpST-1}
    \end{subfigure}
    \begin{subfigure}[b]{0.19\textwidth}
        \includegraphics[width=\textwidth]{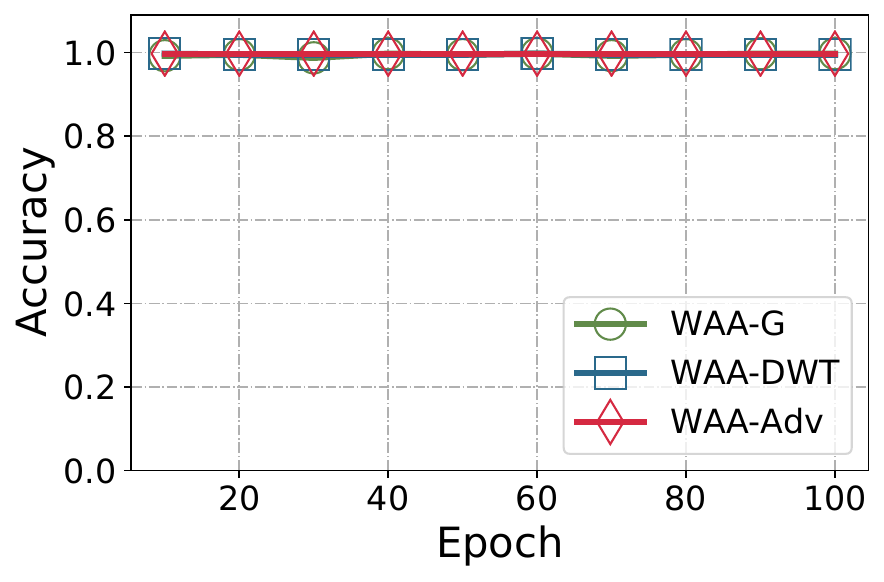}
        \caption{Gaussian Noise}
         \label{fig:gaussianST-1}
    \end{subfigure}
             \begin{subfigure}[b]{0.19\textwidth}
        \includegraphics[width=\textwidth]{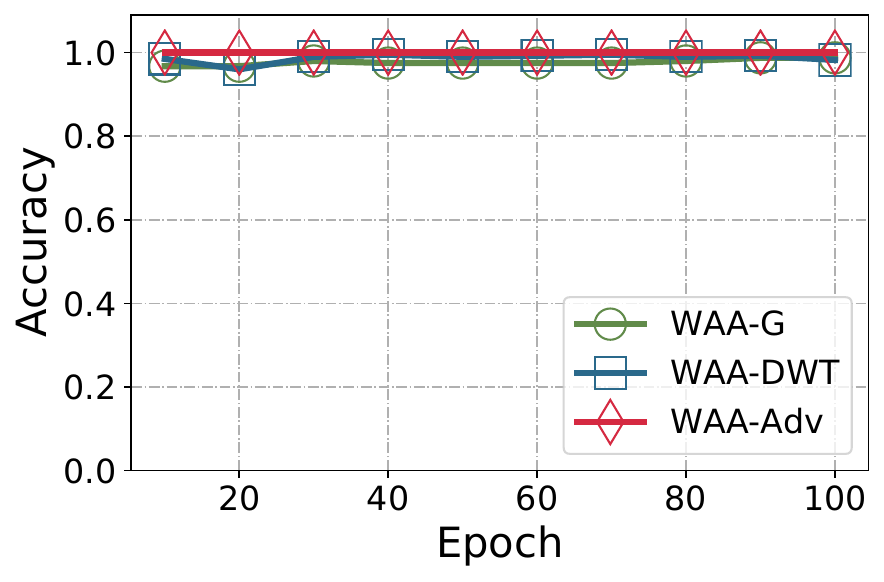}
        \caption{Gaussian Blur}
         \label{fig:blurST-1}
    \end{subfigure}
          \begin{subfigure}[b]{0.19\textwidth}
        \includegraphics[width=\textwidth]{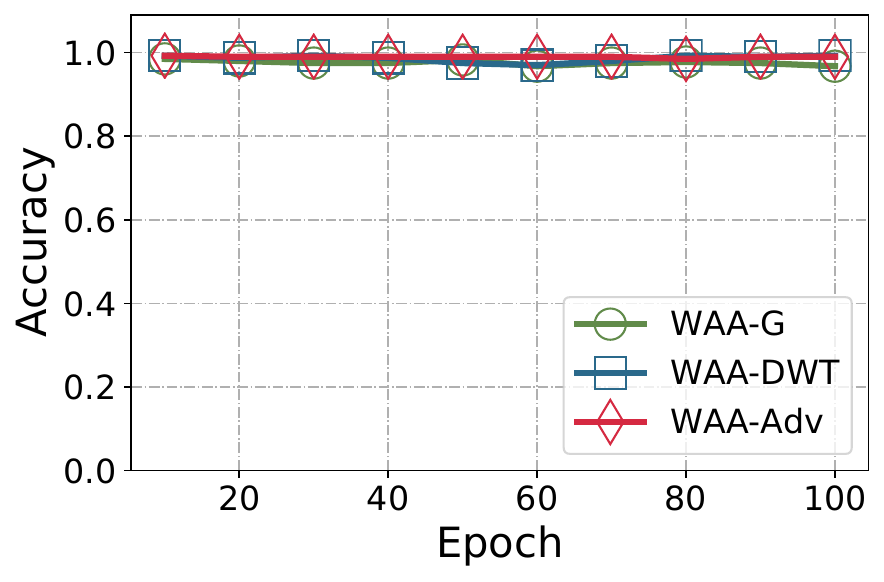}
        \caption{Resizing}
         \label{fig:resizeST-1}
    \end{subfigure}
    \caption{Results on robustness: WikiArt; full parameter fine-tuning}
    \label{fig:robustness:wikiart}
\end{figure*}

\myparagraph{Effectiveness and Image Synthesis Quality} For each combination activation token, we train the corresponding model and use the $10$ activation tokens as prompts to generate $100$ images for each token. We categorize the images generated by each of the $3$ models under its specific activation tokens as category $1$, and those generated by each model under the other prompts as category $2$. We select $1,000$ images to form the training set and 400 images for the test set. For each watermarked method on each dataset, we train a single general classifier to detect whether the generated images contain watermarks. The results are detailed in Figure~\ref{fig:comb-acc}. The classification accuracy of WAA-G, WAA-DWT, and WAA-Adv are 99.5\%/96.8\%, 100\%/99.5\%, and 100\%/100\%, respectively, demonstrating that our approach can effectively handle large-scale requests. The FID results are also detailed in Figure~\ref{fig:comb-fid}.

\myparagraph{Different Noise Budgets for the WAA-Adv Method} Due to the space limit, we put the results of this set of experiments in Appendix~\ref{app:exp:different_noise_budgets}.

\myparagraph{Robustness} For the datasets that can be triggered by the three combination activation tokens mentioned in Section~\ref{sec:traceability}, we explore their robustness with the WAA methods. We assume that the published watermarked images are damaged during the transmission process. In this scenario, we assess the validity of the watermark under different conditions. Five methods are evaluated: JPEG compression, sharpness enhancement, Gaussian noise, Gaussian blur, and resizing. For JPEG compression, we retain only 5\% of the original image quality. For sharpness enhancement, the factor is set to 10. For Gaussian noise, we set the mean to 0 and the variance to 1. For Gaussian blur, the standard deviation is set to 1. For resizing, the image is resized from $512 \times 512$ to $256 \times 256$, and then resized back to the original dimensions.

Afterward, we perform training and generate images. The detection results are detailed in Figure~\ref{fig:robustness:wikiart}. The results show that our method is robust and retains a high detection accuracy in all cases of damage. Due to the space limit, we only present the results on the WikiArt dataset here, and the results on the COCO dataset are included in Appendix~\ref{app:exp:robustness}.

\myparagraph{Multi-User Tracking} As detailed in Section~\ref{sec:details}, for accurate identification of the leaker, it is essential that the image generated by the model for each candidate token is correctly detected by the watermark detector. Following Algorithm~\ref{algorithm-1} and Algorithm~\ref{algorithm-2}, we consider a scenario of multi-user tracking. For 100 users, We successfully track 91 users, achieving a high success rate. The detailed setting and results can be found in Appendix~\ref{app:exp:multiuser}.

\section{Related Work}
\myparagraph{Dataset Protection} A major approach is to add adversarial noise to the image directly such that the generation cannot be completed, e.g.,\cite{liang2023adversarial,van2023anti}. AdvDM~\cite{liang2023adversarial}, based on latent diffusion models, utilizes a Monte-Carlo estimation of adversarial examples for diffusion models by optimizing various latent variables sampled from the reverse process of the models. Anti-DreamBooth~\cite{van2023anti}, which builds on the DreamBooth training method for stable diffusion, introduces subtle noise perturbations to disrupt the generation quality of any DreamBooth model trained on these perturbed images. These efforts effectively disrupted the normal generation process of the model, rendering the produced images unusable and thus preventing malicious individuals from stealing the images. However, this also prevents authorized users from viewing the images, thereby failing to achieve the intended purpose of regular use. Another major method is to add a watermark to the image, e.g.,\cite{swanson1998multimedia,abdelnabi2021adversarial,guan2022deepmih,ma2023generative,ma2023generative,cui2023ft}. In particular, Ma et al. ~\cite{ma2023generative} focus on protecting images generated by subject-driven models, using a method similar to adversarial training to create the watermark generator and detector, and fine-tuning the detector on the generated image set. Cui et al. ~\cite{cui2023ft} claim that their proposed watermarking method is superior to that of Ma et al. ~\cite{ma2023generative}, as it allows the model to learn that the watermark appears before the specific object in the image, thus enhancing protection. These approaches require an impractically high injection ratio for protection and do not allow for identifying the source of a leak.

\myparagraph{Backdoor Attacks and Watermarking}
Backdoor attack~\cite{gu2017badnets} aims to embed hidden malicious behavior in a model during training, which can be activated by specific triggers to cause the model to produce incorrect or harmful outputs. Deep neural networks (DNNs) have been found to be vulnerable to backdoor attacks~\cite{chen2018detecting,liu2018fine,bagdasaryan2020backdoor,zhu2019transferable}. In recent years, with the broader application of text-to-image diffusion models, their security has garnered widespread attention. Some works~\cite{chou2023backdoor,chou2024villandiffusion} inject backdoors into diffusion models by modifying the loss function during the forward and reverse processes, enabling the model to generate images desired by the attacker under specific inputs. Struppek et al.~\cite{struppek2023rickrolling} inject backdoors into pre-trained text encoders. If the text encoders are used in text-to-image diffusion models, the models are injected with a backdoor. These methods effectively inject backdoors into the model but require the attacker to have high capabilities, such as directly training the model.

\section{Concluding Remarks}
This work studies the problem of detecting dataset abuse during the fine-tuning of Stable Diffusion models in the context of text-to-image synthesis. We propose an effective dataset watermarking framework designed to identify unauthorized use and trace the source of any data leaks. Our framework encompasses two main strategies across three different watermarking schemes and extends to scenarios involving large-scale dataset authorization. Through extensive experiments, we demonstrate that our framework is highly effective. Both the TWA and WAA methods require only a minimal injection ratio into the dataset, with WAA, in particular, not requiring any changes to the text, ensuring high stealthiness while preserving the dataset's normal functionality and strong protection effectiveness.

\clearpage
\section*{Ethics Considerations Statement}
This study proposes a framework for detecting dataset abuse in text-to-image synthesis to support copyright and data asset protection. We have thoroughly considered the potential ethical implications of our approach and ensured that our work adheres to established ethical standards. All datasets used in this project are publicly available, and the sources have been properly cited. The data collection process complies with relevant legal and ethical guidelines. As our research does not involve human subjects, no formal Institutional Review Board (IRB) approval is required, per our institution’s guidelines.

\section*{Compliance with the Open Science Policy}
We commit to adhering to the open science policy. The code for this work will be provided in the future version of the paper. 

\bibliographystyle{plain}
\bibliography{bib}

\appendix
\renewcommand{\thesection}{\Alph{section}}

\begin{figure*}[!t]
    \centering
    \begin{subfigure}[b]{0.19\textwidth}
       \includegraphics[width=\textwidth]{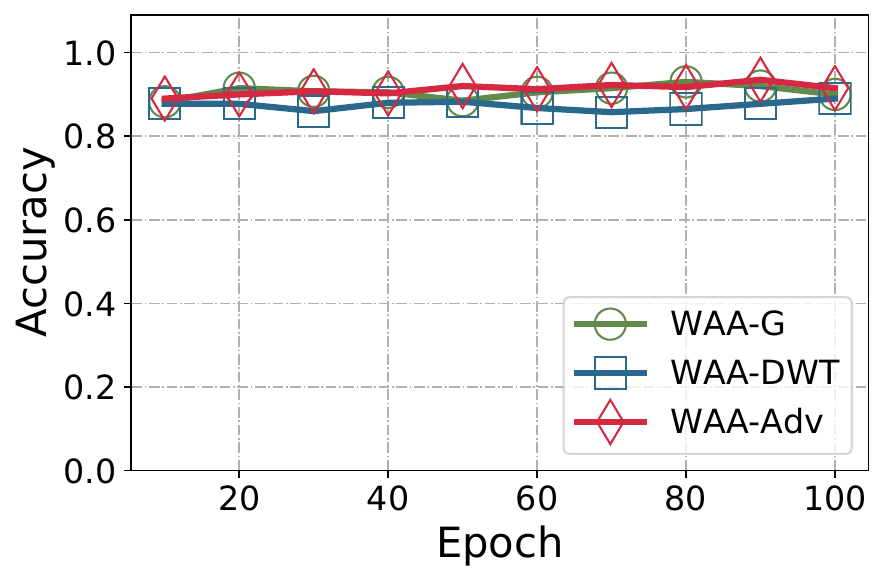}
        \caption{JPEG Compression}
        \label{fig:compST-2}
    \end{subfigure}
     \begin{subfigure}[b]{0.19\textwidth}
        \includegraphics[width=\textwidth]{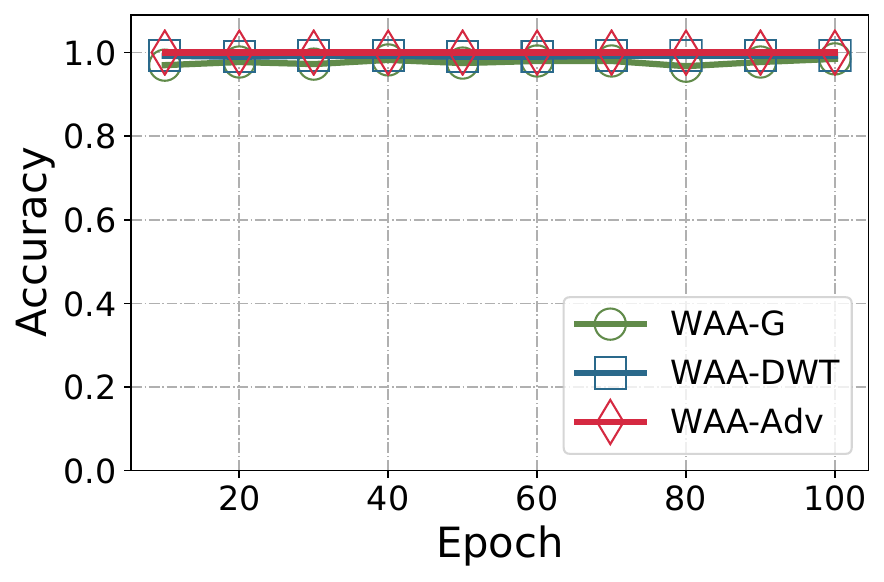}
        \caption{Sharpness}
         \label{fig:sharpST-2}
        \end{subfigure}
     \begin{subfigure}[b]{0.19\textwidth}
        \includegraphics[width=\textwidth]{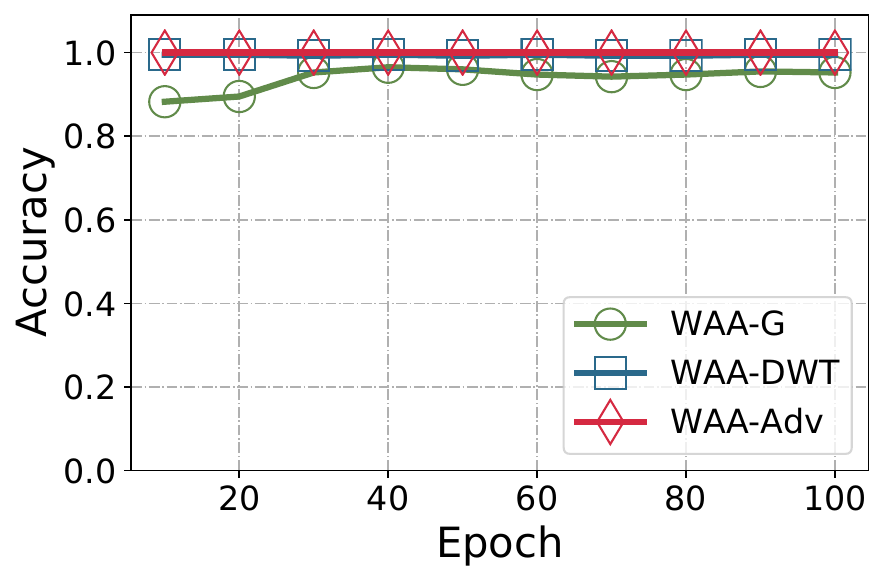}
        \caption{Gaussian Noise}
         \label{fig:gaussianST-2}
        \end{subfigure}
     \begin{subfigure}[b]{0.19\textwidth}
        \includegraphics[width=\textwidth]{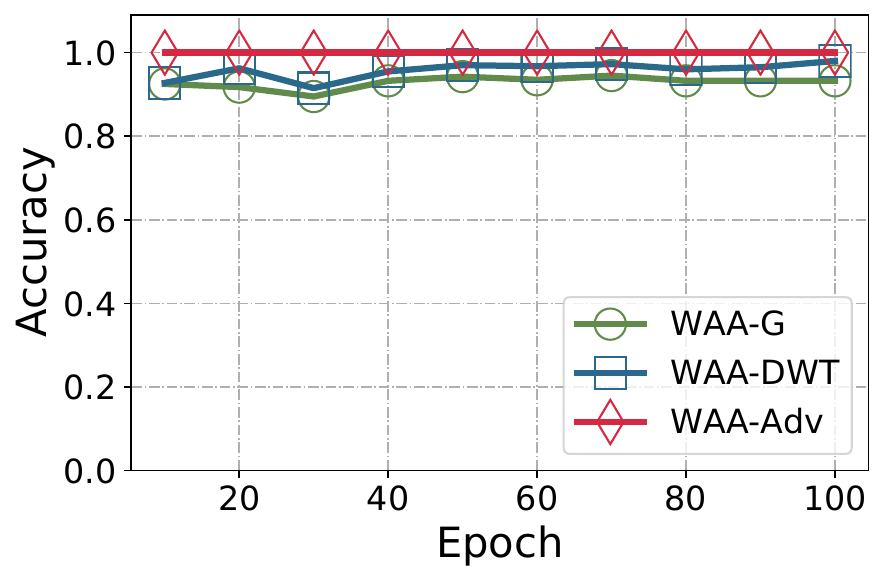}
        \caption{Gaussian Blur}
         \label{fig:blurST-2}
        \end{subfigure}
         \begin{subfigure}[b]{0.19\textwidth}
        \includegraphics[width=\textwidth]{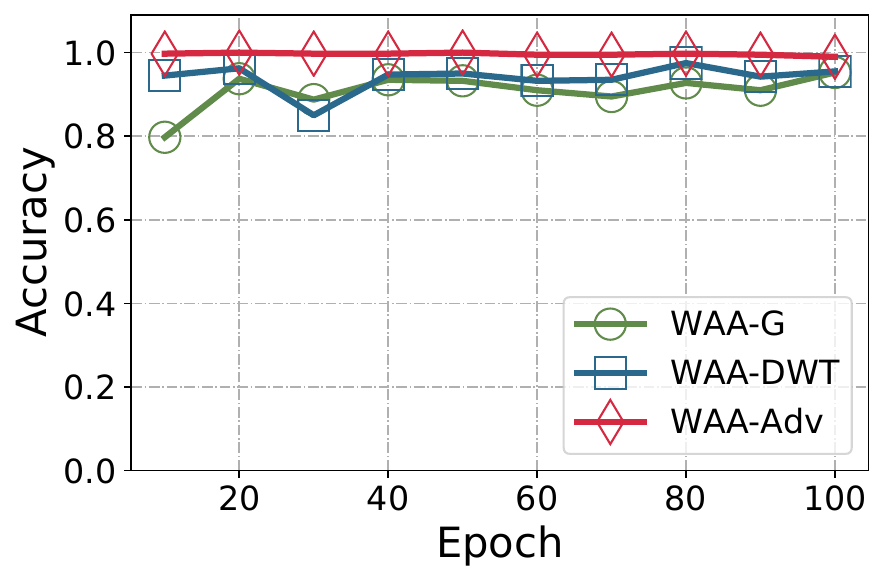}
        \caption{Resizing}
         \label{fig:resizeST-2}
        \end{subfigure}
    \caption{Results on robustness: COCO; full parameter fine-tuning}
    \label{fig:robustness:coco}
\end{figure*}

\section{Appendix}

\subsection{Hyperparameters Settings}
\label{sec:hyperparameters}

In this section, the hyperparameters used in the experiment are provided.

\subsubsection{Fine-Tuning Methods}

\myparagraph{Standard Training} The batch size is set to \(2\) for \(10{,}000\) training steps. The learning rate is \(10^{-4}\). Training a model on a single Nvidia RTX 6000 GPU with 48GB of memory takes approximately 70 minutes.

\myparagraph{LoRA} The batch size is set to \(2\) for \(10{,}000\) training steps. The learning rate is \(10^{-4}\). Training a model on a single Nvidia RTX 6000 GPU with 48GB of memory takes approximately 40 minutes.

\subsubsection{Watermarking Methods}
For the Gaussian method, the mean is set to \(0\), and the standard deviation is set to \(5\).  For the DWT method, \(W\) is created by generating random values, each within the range of \(0\) to \(10\).  For the adversarial method, the noise budget \(\eta\) is set to \(0.05\) (except in Section~\ref{sec:traceability}, which explores the effects under different noise budgets).

\subsection{Additional Experimental Results}
\subsubsection{Results on Robustness: COCO}
\label{app:exp:robustness}
We use the same settings as those applied for evaluating the robustness of the proposed method on the WikiArt dataset. The results, shown in Figure~\ref{fig:robustness:coco}, demonstrate that our method is robust and maintains high classification accuracy across all damage cases on the COCO dataset.

\subsubsection{Different Noise Budgets for the WAA-Adv Method}
\label{app:exp:different_noise_budgets}
\begin{figure}[tb]
  \centering
\begin{subfigure}[b]{0.2\textwidth}
       \includegraphics[width=\textwidth]{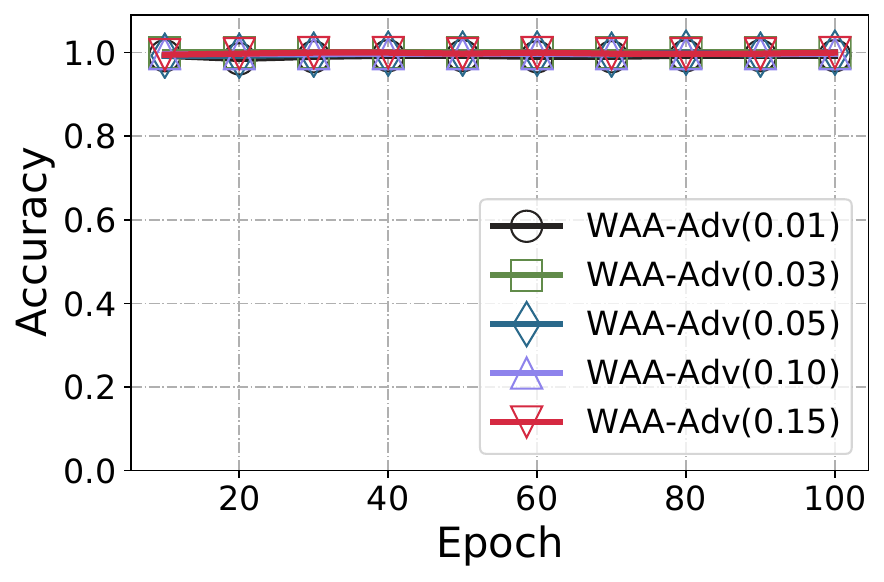}
        \caption{WikiArt, Standard}
    \end{subfigure}
    \begin{subfigure}[b]{0.2\textwidth}
        \includegraphics[width=\textwidth]{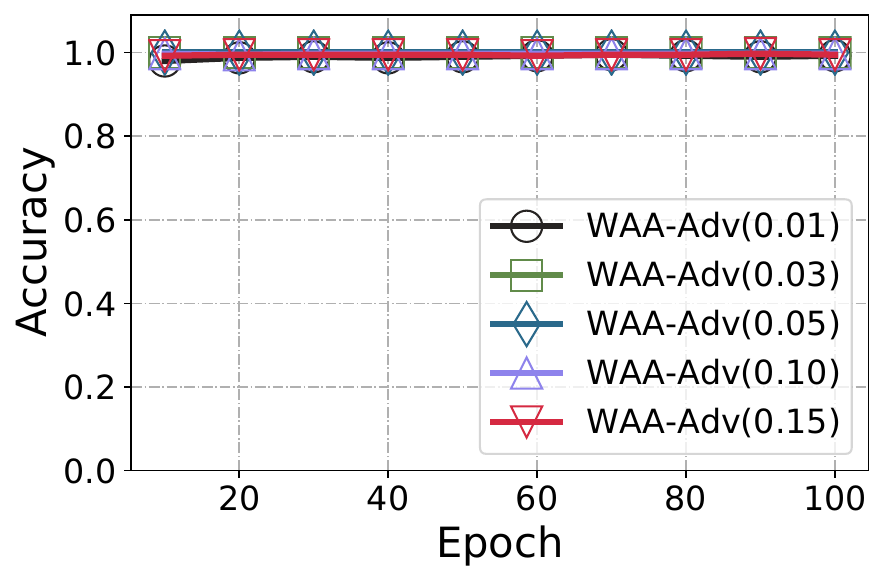}
        \caption{COCO, Standard}
         \label{fig:cat-adv-noise}
    \end{subfigure}
    \caption{Accuracy for the selected combination activation tokens with different noise budgets for the WAA-Adv method.}
    \label{fig:acc-adv-noise}
\end{figure}

\begin{figure}[tb]
  \centering
\begin{subfigure}[b]{0.22\textwidth}
       \includegraphics[width=\textwidth]{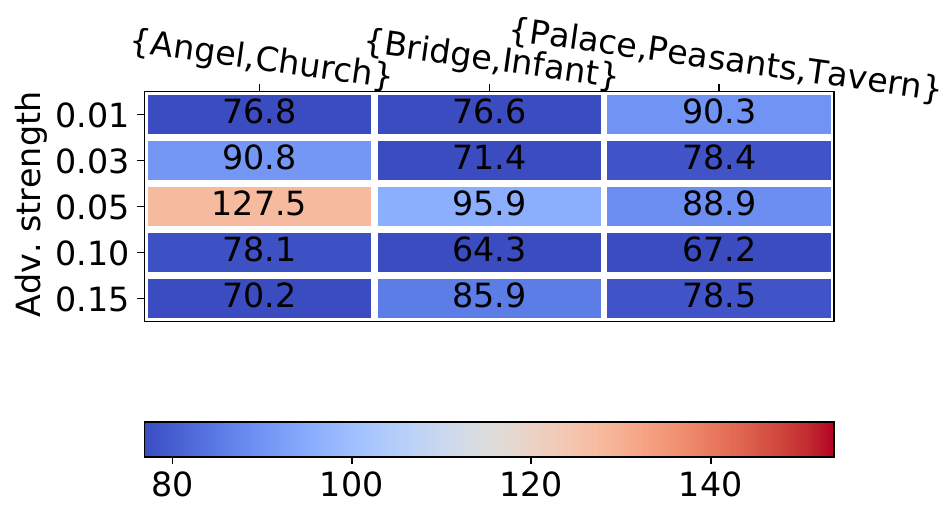}
        \caption{WikiArt, Standard}
    \end{subfigure}
    \begin{subfigure}[b]{0.20\textwidth}
        \includegraphics[width=\textwidth]{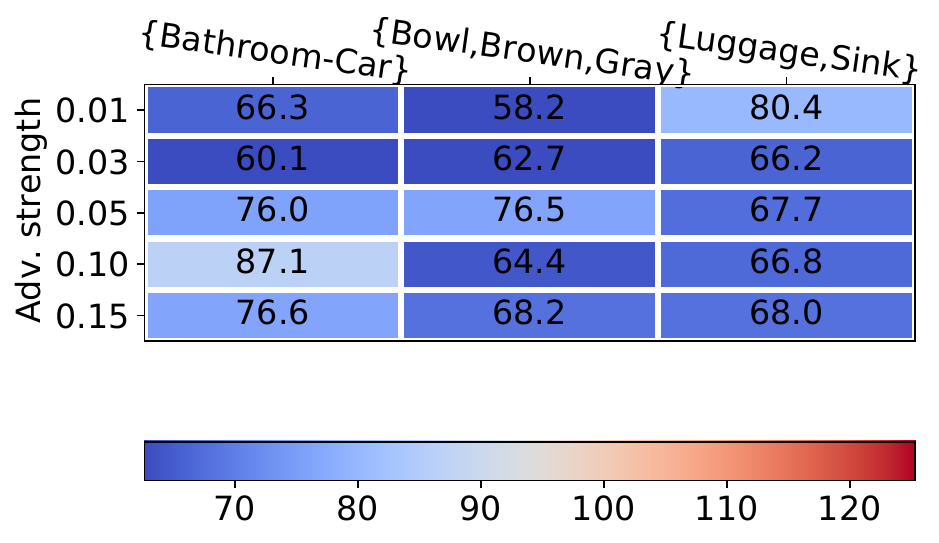}
        \caption{COCO, Standard}
         \label{fig:fid-cat-adv-noise}
    \end{subfigure}
    \caption{FID for the selected combination activation tokens with different noise budgets for the WAA-Adv method.}
    \label{fig:adv-noise-fid}
\end{figure}

We also explore the impact of different noise budgets on image generation for the WAA-Adv method. The noise budgets for adversarial examples are set at 0.01, 0.03, 0.05, 0.10, and 0.15, respectively. For the evaluation of detection accuracy and FID, the number of images generated, as well as the division of training and test sets, remains consistent with the previously described methodology. The detection results and FID results are detailed in Figure~\ref{fig:acc-adv-noise} and Figure~\ref{fig:adv-noise-fid}. The results show that different noise budgets within our selected range have a small impact on the quality of the generated images. Images generated with higher noise budgets exhibit a higher classification accuracy.

\subsubsection{Results on Multi-User Tracking}
\label{app:exp:multiuser}
For the dataset distribution process (Algorithm~\ref{algorithm-1}) using the wikiArt dataset, we set $T=100$, $L=2$, and $R=5$ for dataset generation, with candidate tokens $S$ selected as shown in Table~\ref{table:BackdoorWikiart}. We use the classifier trained for the WAA-Adv method mentioned in Section~\ref{sec:watermarkingeffectiveness} as the classifier.  For each published dataset, we train a substitute model, and for each model, we generate images using tokens in $S$ as detailed in Algorithm~\ref{algorithm-2}. The results are presented in Table~\ref{table:trace-wikiart-1} and Table~\ref{table:trace-wikiart-2}. For $T=100$ users, We successfully track 91 users, achieving a high success rate.

\subsection{Details of Image Watermarking Schemes}

\subsubsection{Gaussian Noise Watermarking}
\label{app:watermarking_methods_gaussian}
Gaussian noise is readily considered as a potential watermarking technique, as it is already utilized in both the forward and reverse processes of stable diffusion. We can sample the noise from a normal distribution and add it to the selected images.
This process involves generating random noise values that conform to a Gaussian distribution and then embedding these values into the images. Formally, the generated Gaussian noise \( z \) follows a normal distribution as $z \sim \mathcal{N}(\mu, \sigma^2)$. For a selected origin image $x$, the watermarked image is $x' = x + z$.

\subsubsection{DWT Watermarking}
\label{app:watermarking_methods_dwt}
The watermarking process using DWT involves three main steps: (1) \textit{Image Decomposition}: For each channel of the selected image $x$, the data owner applies the two-dimensional DWT to obtain the approximation coefficients $cA$ and horizontal, vertical, and diagonal detail coefficients $cH$, $cV$, and $cD$, respectively. (2) \textit{Watermark Embedding}: The watermark is embedded within the horizontal detail coefficients matrix $cH$. A watermark matrix \( W \) is generated with elements randomly chosen and resized to match the dimensions of \( cH \). (3) \textit{Image Reconstruction}: Using the altered coefficients, the Inverse Discrete Wavelet Transform (IDWT) is performed for each channel. These channels are then combined to form the watermarked color image $x'$.

\begin{table}[h!]
\centering
\caption{Validation results for WAA-Adv method.}
\footnotesize
\begin{tabular}{|c|c|c|c|}
\hline
\textbf{User} & \textbf{Prompts} &\textbf{Frequency} & \textbf{Success}\\
\hline
1 & bridge lord charles angel tavern & 0.071 & \faCheck  \\
\hline
2 & bridge church infant & 0.035 & \faCheck \\
\hline
3&church bridge maria tavern &0.060& \faCheck \\
\hline
4&angel church &0.023&  \faCheck \\
\hline
5&lord bridge infant &0.034& \faCheck\\
\hline
6&church peasants lord angel &0.050& \faCheck \\
\hline
7&tavern lord infant &0.036& \faCheck \\
\hline
8&peasants tavern maria palace bridge &0.073& \faCheck \\
\hline
9&palace peasants tavern& 0.040& \faCheck\\
\hline
10&peasants charles church bridge& 0.060& \faCheck \\
\hline
11&maria palace bridge lord & 0.052&\faCheck   \\
\hline
12&bridge tavern angel lord &0.054&\faCheck \\
\hline
13&maria palace &0.028&\faCheck \\
\hline
14&palace bridge  &0.025&\faCheck \\
\hline
15&tavern maria church infant &0.055& \\
\hline
16&maria bridge infant tavern angel &0.073& \faCheck\\
\hline
17&infant maria charles bridge peasants&0.077&\faCheck \\
\hline
18&maria  bridge&0.033&\faCheck \\
\hline
19&maria angel infant charles& 0.057&\faCheck \\
\hline
20&church maria infant palace lord& 0.057&\faCheck \\
\hline
21&tavern peasants church lord &0.049&\faCheck \\
\hline
22&angel tavern palace bridge maria& 0.073& \\
\hline
23&maria peasants palace &0.046& \faCheck\\
\hline
24&tavern church &0.027& \faCheck\\
\hline
25&lord palace angel church bridge&0.057&\\
\hline
26&angel bridge peasants church&0.056&\faCheck\\
\hline
27&peasants bridge infant lord&0.052&\faCheck\\
\hline
28&palace tavern peasants infant lord&0.059&\faCheck\\
\hline
29&bridge tavern lord&0.041&\faCheck\\
\hline
30&maria charles bridge&0.049&\faCheck\\
\hline
31&tavern infant peasants bridge lord&0.064&\faCheck\\
\hline
32&charles peasants infant&0.045&\faCheck\\
\hline
33&lord peasants church&0.037&\\
\hline
34&charles palace tavern lord infant&0.063&\faCheck\\
\hline
35&charles palace bridge&0.042&\faCheck\\
\hline
36&infant tavern church&0.037&\faCheck\\
\hline
37&charles infant&0.027&\faCheck\\
\hline
38&infant angel&0.023&\faCheck\\
\hline
39&bridge maria charles lord angel&0.071&\\
\hline
40&tavern lord palace&0.036&\faCheck\\
\hline
41&tavern charles&0.034&\faCheck\\
\hline
42&church infant &0.020&\faCheck\\
\hline
43&palace peasants lord church infant&0.057&\faCheck\\
\hline
44&maria peasants&0.036&\faCheck\\
\hline
45&angel lord&0.022&\faCheck\\
\hline
46&tavern palace&0.027&\faCheck\\
\hline
47&infant lord&0.019&\faCheck\\
\hline
48&angel charles peasants palace&0.058&\faCheck\\
\hline
49&maria infant&0.028&\faCheck\\
\hline
50&peasants lord charles&0.044&\faCheck\\
\hline
\end{tabular}
\label{table:trace-wikiart-1}
\end{table}

\begin{table}[h!]
\centering
\caption{Validation results for WAA-Adv method} 
\footnotesize
\begin{tabular}{|c|c|c|c|}
\hline
\textbf{user} & \textbf{Prompts} &\textbf{Frequency} & \textbf{Success}\\
\hline
51&palace infant tavern&0.037&\faCheck\\
\hline
52&peasants bridge maria infant lord&0.070&\\
\hline
53&palace maria bridge &0.043&\faCheck\\
\hline
54&maria peasants bridge lord &0.060&\faCheck\\
\hline
55&lord church tavern &0.036&\faCheck\\
\hline
56&lord angel tavern palace &0.049&\faCheck\\
\hline
57&angel maria &0.031&\faCheck\\
\hline
58&angel peasants tavern &0.043&\faCheck\\
\hline
59&infant charles maria &0.044&\faCheck\\
\hline
60&maria infant peasants &0.046&\faCheck\\
\hline
61&lord bridge palace &0.034&\faCheck\\
\hline
62&lord peasants &0.027&\faCheck\\
\hline
63&charles angel lord &0.039&\faCheck\\
\hline
64&angel lord bridge peasants palace &0.065&\\
\hline
65&infant church maria bridge &0.053&\faCheck\\
\hline
66&church infant maria angel tavern &0.068&\faCheck\\
\hline
67&charles lord tavern church &0.053&\faCheck\\
\hline
68&charles peasants &0.035&\faCheck\\
\hline
69&lord angel tavern &0.039&\faCheck\\
\hline
70&angel bridge infant tavern charles &0.072&\\
\hline
71&church angel lord tavern &0.049&\faCheck\\
\hline
72&palace infant &0.020&\faCheck\\
\hline
73&lord palace &0.019&\faCheck\\
\hline
74&tavern angel &0.030&\faCheck\\
\hline
75&maria palace charles &0.044&\faCheck\\
\hline
76&charles palace lord &0.036&\faCheck\\
\hline
77&maria tavern peasants angel church &0.071&\faCheck\\
\hline
78&peasants palace lord infant &0.047&\faCheck\\
\hline
79&lord maria charles church infant &0.063&\faCheck\\
\hline
80&infant maria peasants tavern angel &0.071&\faCheck\\
\hline
81&angel charles bridge peasants &0.063&\faCheck\\
\hline
82&infant angel palace &0.033&\faCheck\\
\hline
83&peasants infant &0.028&\faCheck\\
\hline
84&palace charles &0.027&\faCheck\\
\hline
85&church lord &0.019&\faCheck\\
\hline
86&palace tavern peasants angel&0.053&\faCheck\\
\hline
87&maria bridge church palace lord &0.062&\faCheck\\
\hline
88&tavern peasants &0.030&\faCheck\\
\hline
89&infant charles bridge palace& 0.052&\faCheck\\
\hline
90&charles lord &0.026&\faCheck\\
\hline
91&maria charles peasants lord palace &0.071&\faCheck\\
\hline
92&angel infant peasants tavern &0.053&\faCheck \\
\hline
93&infant lord church &0.029&\faCheck\\
\hline
94&angel peasants &0.031&\faCheck\\
\hline
95&peasants palace charles maria &0.062&\faCheck\\
\hline
96&angel lord infant charles palace &0.059&\\
\hline
97&lord tavern &0.026&\faCheck\\
\hline
98&charles church &0.027&\faCheck\\
\hline
99&bridge tavern &0.032&\faCheck\\
\hline
100&tavern palace angel peasants bridge& 0.068&\faCheck\\
\hline
\end{tabular}
\label{table:trace-wikiart-2}
\end{table}

\clearpage

\subsection{Algorithms for Large-Scale Authorization with WAA methods}
This part presents the algorithms for large-scale authorization with WAA methods. The pseudocodes of algorithms for distributing datasets and detecting the data leak are shown in Algorithm~\ref{algorithm-1} and Algorithm~\ref{algorithm-2}, respectively.

\begin{algorithm}
\caption{Distribute Datasets to Users for WAA methods}
\label{algorithm-1}
\textbf{Input:} Number of data users $T$; Dataset $D=\{(image_i, text_i)\}$; Set of candidate tokens $S$; Minimum number of tokens $L$; Maximum number of tokens $R$.
\begin{algorithmic}[1]
 \STATE Initialize $V_{\text{all}} \gets \{\}$ \COMMENT{Set to store all selected $V_t$}
 \FOR{$t = 1$ to $T$}
    \REPEAT
        \STATE  $V_t\gets$ Randomly select $L \sim R$ tokens from $S$
    \UNTIL{$V_t \notin V_{\text{all}}$}
    \STATE Add $V_t$ to $V_{\text{all}}$
    \STATE Initialize $D_t \gets \{\}$
    \FOR {$i = 1$ to $|D|$}
        \STATE Extract $words_i$ as the set of words in $text_i$
        \STATE Initialize $newImage_i \gets image_i$
        \IF{$V_t \cap words_i \neq \emptyset$}
            \STATE Apply corresponding watermarking $w_i$ to $x_i$
            \STATE Update $newImage_i \gets newImage_i + w_i$
        \ENDIF
        \STATE Add $(newImage_i, text_i)$ to $D_t$
    \ENDFOR
    \STATE Distribute $D_t$ to user $t$
  \ENDFOR
\end{algorithmic}
\end{algorithm}

\begin{algorithm}
\caption{Traceable Detection for WAA methods}
\label{algorithm-2}
\textbf{Input:} Target diffusion model $\mathcal{G}$; Pretrained binary classification model $\mathcal{D}$; The set of candidate tokens $S$; The number of data users $T$; Selected tokens for user $t$ $V_t$.
\begin{algorithmic}[1]
\STATE Initialize $W \gets \{\}$
 \FOR{$s \in S$}
    \STATE $p\gets$ generate image with $\mathcal{G}$ using prompt $s$
    \IF{$\mathcal{D}$ determines that $p$ has watermark}
        \STATE Add $s$ to $W$
    \ENDIF
 \ENDFOR
 \FOR{$t = 1$ to $T$}
    \IF{$equal(W, V_t)$}
        \STATE \textbf{output} $\mathcal{D}$ generated by the data distributed to user t
        \RETURN
    \ENDIF
 \ENDFOR
\end{algorithmic}
\end{algorithm}


\subsection{Token Selection}

\label{sec:AppendTokenSelect}
In this section, we list the activation tokens selected for each method in our experiments. Table~\ref{table:BackdoorWikiart} shows the selected activation tokens in the WAA methods with the standard fine-tuning and LoRA methods on the WikiArt dataset. Table~\ref{table:BackdoorCOCO} shows the selected activation tokens in the WAA methods with the standard fine-tuning method and LoRA method on the COCO dataset. Table~\ref{table:BackdoorComb} shows the selected activation tokens for experiments on large-scale authorization.

\begin{table*}[h!]
\centering
\caption{The selected token and its frequency across all 1000 images in the WikiArt dataset for the WAA methods.}
\footnotesize
\begin{tabular}{|c|c|c|c|c|}
\hline
\textbf{Index} & \textbf{WikiArt (Standard)} & \textbf{Frequency} & \textbf{WikiArt (LoRA)} & \textbf{Frequency} \\
\hline
1 & angel & 0.013 & child & 0.030 \\
\hline
2 & bridge & 0.016 & family & 0.022 \\
\hline
3 & charles & 0.017 & immaculate & 0.014 \\
\hline
4 & church & 0.010 & inn & 0.016 \\
\hline
5 & infant & 0.010 & interior & 0.016 \\
\hline
6 & lord & 0.009 & lady & 0.015 \\
\hline
7 & maria & 0.018 & landscape & 0.021 \\
\hline
8 & palace & 0.010 & peasant & 0.017 \\
\hline
9 & peasants & 0.018 & self & 0.015 \\
\hline
10 & tavern & 0.017 & virgin & 0.030 \\
\hline
\end{tabular}
\label{table:BackdoorWikiart}
\end{table*}

\begin{table*}[h!]
\centering
\caption{The selected token and its frequency across all 1000 images in the COCO dataset for the WAA methods.}
\footnotesize
\begin{tabular}{|c|c|c|c|c|}
\hline
\textbf{Index} & \textbf{COCO (Standard)} & \textbf{Frequency} & \textbf{COCO (LoRA)} & \textbf{Frequency} \\
\hline
1 & bathroom & 0.030 & bathroom & 0.030 \\
\hline
2 & bowl & 0.022 & car & 0.032 \\
\hline
3 & brown & 0.025 & keyboard & 0.019 \\
\hline
4 & car & 0.032 & luggage & 0.020 \\
\hline
5 & gray & 0.023 & room & 0.039 \\
\hline
6 & luggage & 0.020 & shoes & 0.019 \\
\hline
7 & sink & 0.030 & sink & 0.030 \\
\hline
8 & toilet & 0.021 & toilet & 0.021 \\
\hline
9 & tv & 0.023 & tv & 0.023 \\
\hline
10 & under & 0.026 & under & 0.026 \\
\hline
\end{tabular}
\label{table:BackdoorCOCO}
\end{table*}

\begin{table*}[h!]
\centering
\caption{The selected tokens and their total frequency across all 1000 images for combination activation token experiment.}
\footnotesize
\begin{tabular}{|c|c|c|c|c|}
\hline
\textbf{Index} & \textbf{WikiArt} & \textbf{Total Frequency} & \textbf{COCO} & \textbf{Total Frequency} \\
\hline
1 & angel, church & 0.023 & bathroom, car & 0.063 \\
\hline
2 & bridge, infant & 0.025 & luggage, sink& 0.051 \\
\hline
3 & palace, peasants, tavern & 0.040 & bowl, brown, gray & 0.071 \\
\hline
\end{tabular}
\label{table:BackdoorComb}
\end{table*}

\end{document}